\documentclass{article}

\usepackage{microtype}
\usepackage{graphicx}
\usepackage{subfigure}
\usepackage{booktabs} 

\usepackage{hyperref}
\usepackage{multirow}
\usepackage{xspace}

\usepackage{dblfloatfix}
\usepackage{transparent}
\usepackage{algorithm}
\usepackage{listings}
\usepackage{amsmath}
\usepackage{amsthm}
\usepackage{amsfonts}
\usepackage{bm}
\usepackage{amssymb}
\usepackage{bbm}
\usepackage{mathtools}
\usepackage{enumitem}
\usepackage{nicefrac}

\usepackage[accepted]{icml2020}

\begin{document}

\newcommand{\alg}{\ensuremath{{\rm STR}}\xspace}

\newtheorem{theorem}{Theorem}[section]
\newtheorem{corollary}{Corollary}[theorem]
\newtheorem{lemma}[theorem]{Lemma}
\newtheorem{proposition}[theorem]{Proposition}

\newtheorem{definition}[theorem]{Definition}
\newtheorem{remark}[theorem]{Remark}

\newcommand{\round}[1]{\left( #1 \right)}
\newcommand{\curly}[1]{\left\lbrace #1 \right\rbrace}
\newcommand{\squarebrack}[1]{\left\lbrack #1 \right\rbrack}

\newcommand{\sumi}[2]{\sum\limits_{i=#1}^{#2}}
\newcommand{\sumj}[2]{\sum\limits_{j=#1}^{#2}}
\newcommand{\sumk}[2]{\sum\limits_{k=#1}^{#2}}
\newcommand{\sump}[2]{\sum\limits_{p=#1}^{#2}}
\newcommand{\suml}[2]{\sum\limits_{l=#1}^{#2}}
\newcommand{\sumn}[2]{\sum\limits_{n=#1}^{#2}}
\newcommand{\summ}[2]{\sum\limits_{m=#1}^{#2}}
\newcommand{\sumt}[2]{\sum\limits_{t=#1}^{#2}}

\newcommand{\Sum}{\sum_{i = 1}^{n}}
\newcommand{\Sumi}[1]{\sum\limits_{i = 1}^{#1}}
\newcommand{\Sumt}[1]{\sum\limits_{t = 1}^{#1}}

\newcommand{\abs}[1]{\left\lvert #1 \right\rvert}
\newcommand{\norm}[2]{\left\lVert#2\right\rVert_{#1}}
\newcommand{\esqnorm}[1]{\left\lVert#1\right\rVert_2^2}
\newcommand{\enorm}[1]{\left\lVert#1\right\rVert_2}
\newcommand{\infnorm}[1]{\left\lVert#1\right\rVert_\infty}
\newcommand{\opnorm}[1]{\left\lVert#1\right\rVert_\text{op}}
\newcommand{\normF}[1]{\left\lVert#1\right\rVert_{\text{F}}}
\newcommand{\inner}[1]{\left\langle#1\right\rangle}
\newcommand{\ceil}[1]{\left\lceil#1\right\rceil}
\newcommand{\floor}[1]{\left\lfloor#1\right\rfloor}
\newcommand{\sign}[1]{\mathrm{sign}\left(#1\right)}
\newcommand{\supp}[1]{\mathrm{supp}\left(#1\right)}

\newcommand{\zero}{\mathbf{0}}
\newcommand{\one}{\mathbf{1}}

\newcommand{\avec}{\mathbf{a}}
\newcommand{\bvec}{\mathbf{b}}
\newcommand{\cvec}{\mathbf{c}}
\newcommand{\dvec}{\mathbf{d}}
\newcommand{\e}{\mathbf{e}}
\newcommand{\f}{\mathbf{f}}
\newcommand{\g}{\mathbf{g}}
\newcommand{\h}{\mathbf{h}}
\newcommand{\ivec}{\mathbf{i}}
\newcommand{\jvec}{\mathbf{j}}
\newcommand{\kvec}{\mathbf{k}}
\newcommand{\lvec}{\mathbf{l}}
\newcommand{\m}{\mathbf{m}}
\newcommand{\n}{\mathbf{n}}
\newcommand{\ovec}{\mathbf{o}}
\newcommand{\p}{\mathbf{p}}
\newcommand{\q}{\mathbf{q}}
\newcommand{\rvec}{\mathbf{r}}
\newcommand{\s}{\mathbf{s}}
\newcommand{\tvec}{\mathbf{t}}
\newcommand{\uvec}{\mathbf{u}}
\newcommand{\vvec}{\mathbf{v}}
\newcommand{\w}{\mathbf{w}}
\newcommand{\x}{\mathbf{x}}
\newcommand{\y}{\mathbf{y}}
\newcommand{\z}{\mathbf{z}}

\newcommand{\A}{\mathbf{A}}
\newcommand{\B}{\mathbf{B}}
\newcommand{\C}{\mathbf{C}}
\newcommand{\D}{\mathbf{D}}
\newcommand{\Emat}{\mathbf{E}}
\newcommand{\F}{\mathbf{F}}
\newcommand{\G}{\mathbf{G}}
\newcommand{\Hmat}{\mathbf{H}}
\newcommand{\I}{\mathbf{I}}
\newcommand{\J}{\mathbf{J}}
\newcommand{\K}{\mathbf{K}}
\newcommand{\Lmat}{\mathbf{L}}
\newcommand{\M}{\mathbf{M}}
\newcommand{\N}{\mathbf{N}}
\newcommand{\Omat}{\mathbf{O}}
\newcommand{\Pmat}{\mathbf{P}}
\newcommand{\Q}{\mathbf{Q}}
\newcommand{\Rmat}{\mathbf{R}}
\newcommand{\Smat}{\mathbf{S}}
\newcommand{\T}{\mathbf{T}}
\newcommand{\U}{\mathbf{U}}
\newcommand{\V}{\mathbf{V}}
\newcommand{\W}{\mathbf{W}}
\newcommand{\X}{\mathbf{X}}
\newcommand{\Y}{\mathbf{Y}}
\newcommand{\Z}{\mathbf{Z}}

\newcommand{\SIGMA}{\mathbf{\Sigma}}
\newcommand{\LAMBDA}{\mathbf{\Lambda}}

\newcommand{\Acal}{\mathcal{A}}
\newcommand{\Bcal}{\mathcal{B}}
\newcommand{\Ccal}{\mathcal{C}}
\newcommand{\Dcal}{\mathcal{D}}
\newcommand{\Ecal}{\mathcal{E}}
\newcommand{\Fcal}{\mathcal{F}}
\newcommand{\Gcal}{\mathcal{G}}
\newcommand{\Hcal}{\mathcal{H}}
\newcommand{\Ical}{\mathcal{I}}
\newcommand{\Jcal}{\mathcal{J}}
\newcommand{\Kcal}{\mathcal{K}}
\newcommand{\Lcal}{\mathcal{L}}
\newcommand{\Mcal}{\mathcal{M}}
\newcommand{\Ncal}{\mathcal{N}}
\newcommand{\Ocal}{\mathcal{O}}
\newcommand{\Pcal}{\mathcal{P}}
\newcommand{\Qcal}{\mathcal{Q}}
\newcommand{\Rcal}{\mathcal{R}}
\newcommand{\Scal}{\mathcal{S}}
\newcommand{\Tcal}{\mathcal{T}}
\newcommand{\Ucal}{\mathcal{U}}
\newcommand{\Vcal}{\mathcal{V}}
\newcommand{\Wcal}{\mathcal{W}}
\newcommand{\Xcal}{\mathcal{X}}
\newcommand{\Ycal}{\mathcal{Y}}
\newcommand{\Zcal}{\mathcal{Z}}

\newcommand{\alphavec}{\boldsymbol{\alpha}}
\newcommand{\betavec}{\boldsymbol{\beta}}
\newcommand{\gammavec}{\boldsymbol{\gamma}}
\newcommand{\deltavec}{\boldsymbol{\delta}}
\newcommand{\epsvec}{\boldsymbol{\epsilon}}
\newcommand{\etavec}{\boldsymbol{\eta}}
\newcommand{\nuvec}{\boldsymbol{\nu}}
\newcommand{\tauvec}{\boldsymbol{\tau}}
\newcommand{\rhovec}{\boldsymbol{\rho}}
\newcommand{\lmbda}{\boldsymbol{\lambda}}
\newcommand{\muvec}{\boldsymbol{\mu}}
\newcommand{\thetavec}{\boldsymbol{\theta}}

\newcommand{\BigO}[1]{\mathcal{O}\round{#1}}
\newcommand{\BigOmega}[1]{\Omega\round{#1}}

\newcommand{\R}{\mathbb{R}}
\newcommand{\Rd}[1]{\mathbb{R}^{#1}}
\newcommand{\Natural}{\mathbb{N}}
\newcommand{\Complex}{\mathbb{C}}
\newcommand{\Integer}{\mathbb{Z}}
\newcommand{\Rational}{\mathbb{Q}}

\newcommand{\E}[1]{\mathbb{E}\squarebrack{#1}}
\newcommand{\Exp}[2]{\mathbb{E}_{#1}\squarebrack{#2}}
\newcommand{\Prob}[1]{P\curly{#1}}
\newcommand{\Var}[1]{\mathrm{Var}\squarebrack{#1}}

\newcommand{\inv}[1]{\frac{1}{#1}}
\newcommand{\indicator}[1]{\mathbf{1}\curly{#1}}
\newcommand{\Tr}[1]{\text{Tr}\squarebrack{#1}}

\newcommand{\BOX}[1]{\fbox{\parbox{\linewidth}{\centering#1}}}
\newcommand{\textequal}[1]{\stackrel{#1}{=}}
\newcommand{\textleq}[1]{\stackrel{#1}{\leq}}
\newcommand{\textgeq}[1]{\stackrel{#1}{\geq}}
\newcommand{\defeq}{\vcentcolon=}

\newcommand{\dd}[2]{\frac{d #1}{d #2}}
\newcommand{\ddn}[3]{\frac{d^{#1} #2}{d #3^{#1}}}
\newcommand{\dodo}[2]{\frac{\partial #1}{\partial #2}}
\newcommand{\dodon}[3]{\frac{\partial^{#1} #2}{\partial {#3}^{#1}}}


\twocolumn[
\icmltitlerunning{Soft Threshold Weight Reparameterization for Learnable Sparsity}

\icmltitle{Soft Threshold Weight Reparameterization for Learnable Sparsity}



\icmlsetsymbol{equal}{*}

\begin{icmlauthorlist}
\icmlauthor{Aditya Kusupati}{uw}\\
\icmlauthor{Vivek Ramanujan}{equal,ai2}
\icmlauthor{Raghav Somani}{equal,uw}
\icmlauthor{Mitchell Wortsman}{equal,uw}\\
\icmlauthor{Prateek Jain}{msr}
\icmlauthor{Sham Kakade}{uw}
\icmlauthor{Ali Farhadi}{uw}
\end{icmlauthorlist}

\icmlaffiliation{msr}{Microsoft Research, India}
\icmlaffiliation{uw}{University of Washington, USA}
\icmlaffiliation{ai2}{Allen Institute for Artificial Intelligence, USA}

\icmlcorrespondingauthor{Aditya Kusupati}{kusupati@cs.washington.edu}

\icmlkeywords{Machine Learning, ICML}

\vskip 0.3in
]



\printAffiliationsAndNotice{\icmlEqualContribution} 

\begin{abstract}

Sparsity in Deep Neural Networks (DNNs) is studied extensively with the focus of maximizing prediction accuracy given an overall parameter budget. Existing methods rely on uniform or heuristic non-uniform sparsity budgets which have sub-optimal layer-wise parameter allocation resulting in a) lower prediction accuracy or b) higher inference cost (FLOPs). This work proposes Soft Threshold Reparameterization (\alg), a novel use of the soft-threshold operator on DNN weights. \alg smoothly induces sparsity while \emph{learning} pruning thresholds thereby obtaining a non-uniform sparsity budget. Our method achieves state-of-the-art accuracy for unstructured sparsity in CNNs (ResNet50 and MobileNetV1 on ImageNet-1K), and, additionally, learns non-uniform budgets that empirically reduce the FLOPs by up to 50\%. Notably, \alg boosts the accuracy over existing results by up to 10\% in the ultra sparse (99\%) regime and can also be used to induce low-rank (structured sparsity) in RNNs. In short, \alg is a simple mechanism which learns effective sparsity budgets that contrast with popular heuristics. Code, pretrained models and sparsity budgets are at \url{https://github.com/RAIVNLab/STR}.

\end{abstract}
\section{Introduction}
\label{sec:intro}
Deep Neural Networks (DNNs) are the state-of-the-art models for many important tasks in the domains of Computer Vision, Natural Language Processing, etc. To enable highly accurate solutions, DNNs require large model sizes resulting in huge inference costs, which many times become the main bottleneck in the real-world deployment of the solutions. During inference, a typical DNN model stresses the following aspects of the compute environment: 1) RAM - working memory, 2) Processor compute - Floating Point Operations (FLOPs\footnote{One Multiply-Add is counted as one FLOP}), and 3) Flash - model size. Various techniques are proposed to make DNNs efficient including model pruning (sparsity)~\citep{han2015learning}, knowledge distillation~\citep{bucilua2006model}, model architectures~\citep{howard2017mobilenets} and quantization~\citep{rastegari2016xnor}.

Sparsity of the model, in particular, has potential for impact across a variety of inference settings as it reduces the model size and inference cost (FLOPs) without significant change in training pipelines. Naturally, several interesting projects address inference speed-ups via sparsity on existing frameworks~\citep{liu2015sparse, elsen2019fast} and commodity hardware~\citep{ashbyexploiting}. On-premise or Edge computing is another domain where sparse DNNs have potential for deep impact as it is governed by billions of battery limited devices with single-core CPUs. These devices, including mobile phones~\citep{anguita2012} and IoT sensors~\citep{patil2019gesturepod, roy2019one}, can benefit significantly from sparsity as it can enable real-time on-device solutions.

Sparsity in DNNs, surveyed extensively in Section~\ref{sec:rw}, has been the subject of several papers where new algorithms are designed to obtain models with a given parameter budget. But state-of-the-art DNN models tend to have a large number of layers with highly non-uniform distribution both in terms of the number of parameters as well as FLOPs required per layer. Most existing methods rely either on uniform sparsity across all parameter tensors (layers) or on heuristic non-uniform sparsity budgets leading to a sub-optimal weight allocation across layers and can lead to a significant loss in accuracy.  Furthermore, if the budget is set at a global level, some of the layers with a small number of parameters would be fully dense as their contribution to the budget is insignificant. However, those layers can have significant FLOPs, e.g., in an initial convolution layer, a simple tiny 3$\times$3 kernel would be applied to the entire image. Hence, while such models might decrease the number of non-zeroes significantly, their FLOPs could still be large. 

Motivated by the above-mentioned challenges, this works addresses the following question: ``{\em Can we design a method to learn non-uniform sparsity budget across layers that is optimized per-layer, is stable, and is accurate?}".

Most existing methods for learning sparse DNNs have their roots in the long celebrated literature of high-dimension statistics and, in particular, sparse regression. These methods are mostly based on well-known Hard and Soft Thresholding techniques, which are essentially projected gradient methods with explicit projection onto the set of sparse parameters. However, these methods require a priori knowledge of sparsity, and as mentioned above, mostly heuristic methods are used to set the sparsity levels per layer. 

We propose Soft Threshold Reparameterization (\alg) to address the aforementioned issues. We use the fact that the projection onto the sparse sets is available in closed form and propose a novel reparameterization of the problem. That is, for forward pass of DNN, we use  soft-thresholded version~\citep{donoho1995noising} of a weight tensor $\mathbf{W}_l$ of the $l$-th layer in the DNN: $\Scal(\mathbf{W}_l,\alpha_l) := \sign{\mathbf{W}_l}\cdot\mathrm{ReLU}(\abs{\mathbf{W}_l}-\alpha_l)$ where  $\alpha_l$ is the pruning threshold for the $l$-th layer. As the DNN loss can be written as a continuous function of $\alpha_l$'s, we can use backpropagation to learn layer-specific $\alpha_l$ to smoothly induce sparsity. Typically, each layer in a neural network is distinct unlike the interchangeable weights and neurons making it interesting to learn layer-wise sparsity.

Due to layer-specific thresholds and sparsity, \alg is able to achieve state-of-the-art accuracy for unstructured sparsity in CNNs across various sparsity regimes. \alg makes even small-parameter layers sparse resulting in models with significantly lower inference FLOPs than the baselines. For example, \alg for 90\% sparse MobileNetV1 on ImageNet-1K results in a 0.3\% boost in accuracy with 50\% fewer FLOPs. Empirically, \alg's learnt non-uniform budget makes it a very effective choice for ultra (99\%) sparse ResNet50 as well where it is $\sim$10\% more accurate than baselines on ImageNet-1K. \alg can also be trivially modified to induce structured sparsity, demonstrating its generalizability to a variety of DNN architectures across domains. Finally, \alg's learnt non-uniform sparsity budget transfers across tasks thus discovering an efficient sparse backbone of the model. 

The 3 major contributions of this paper are:\vspace{-3mm}
\begin{itemize}[leftmargin=*]
    \itemsep 0pt
    \topsep 0pt
    \parskip 1pt
    \item Soft Threshold Reparameterization (\alg), for the weights in DNNs, to induce sparsity via learning the per-layer pruning thresholds thereby obtaining a better non-uniform sparsity budget across layers. 
    \item Extensive experimentation showing that \alg achieves the state-of-the-art accuracy for sparse CNNs (ResNet50 and MobileNetV1 on ImageNet-1K) along with a significant reduction in inference FLOPs. 
    \item Extension of \alg to structured sparsity, that is useful for the direct implementation of fast inference in practice.
\end{itemize}
\section{Related Work}
\label{sec:rw}
This section covers the spectrum of work on sparsity in DNNs. The sparsity in the discussion can be characterized as (a) unstructured and (b) structured while sparsification techniques can be (i) dense-to-sparse, and (ii) sparse-to-sparse. Finally, the sparsity budget in DNNs can either be (a) uniform, or (b) non-uniform across layers. This will be a key focus of this paper, as different budgets result in different inference compute costs as measured by FLOPs. This section also discusses the recent work on learnable~sparsity.

\subsection{Unstructured and Structured Sparsity}
Unstructured sparsity does not take the structure of the model (e.g. channels, rank, etc.,) into account. Typically, unstructured sparsity is induced in DNNs by making the parameter tensors sparse directly based on heuristics (e.g. weight magnitude) thereby creating sparse tensors that might not be capable of leveraging the speed-ups provided by commodity hardware during training and inference. Unstructured sparsity has been extensively studied and includes methods which use gradient, momentum, and Hessian based heuristics~\citep{evci2019rigging, lee2018snip, lecun1990optimal, hassibi1993second,dettmers2019sparse}, and magnitude-based pruning~\citep{han2015learning, guo2016dynamic, zhu2017prune, frankle2018the, gale2019state, mostafa2019parameter, bellec2018deep, mocanu2018scalable, narang2017exploring,kusupati2018fastgrnn,wortsman2019discovering}. Unstructured sparsity can also be induced by $L_0, L_1$ regularization~\citep{louizos2018learning}, and Variational Dropout (VD)~\citep{molchanov2017variational}. 

Gradual Magnitude Pruning (GMP), proposed in ~\citep{zhu2017prune}, and studied further in ~\citep{gale2019state}, is a simple magnitude-based weight pruning applied gradually over the course of the training. Discovering Neural Wirings (DNW)~\citep{wortsman2019discovering} also relies on magnitude-based pruning while utilizing a straight-through estimator for the backward pass. GMP and DNW are the state-of-the-art for unstructured pruning in DNNs (especially in CNNs) demonstrating the effectiveness of magnitude pruning. VD gets accuracy comparable to GMP~\citep{gale2019state} for CNNs but at a cost of $2\times$ memory and $4\times$ compute during training making it hard to be used ubiquitously.

Structured sparsity takes structure into account making the models scalable on commodity hardware with the standard computation techniques/architectures. Structured sparsity includes methods which make parameter tensors low-rank~\citep{jaderberg2014speeding, alizadeh2019butterfly,lu2016learning}, prune out channels,  filters and induce block/group sparsity~\citep{liu2018rethinking, wen2016learning,li2016pruning, luo2017thinet,gordon2018morphnet,yu2019network}. Even though structured sparsity can leverage speed-ups provided by parallelization, the highest levels of model pruning are only possible with unstructured sparsity techniques.

\subsection{Dense-to-sparse and Sparse-to-sparse Training}
Until recently, most sparsification methods were dense-to-sparse i.e., the DNN starts fully dense and is made sparse by the end of the training. Dense-to-sparse training in DNNs encompasses the techniques presented in~\citep{han2015learning, zhu2017prune, molchanov2017variational, frankle2018the, renda2020comparing}.

The lottery ticket hypothesis~\citep{frankle2018the} sparked an interest in training sparse neural networks end-to-end. This is referred to as sparse-to-sparse training and a lot of recent work~\citep{mostafa2019parameter, bellec2018deep, evci2019rigging, lee2018snip, dettmers2019sparse} aims to do sparse-to-sparse training using techniques which include re-allocation of weights to improve accuracy. 

Dynamic Sparse Reparameterization (DSR)~\citep{mostafa2019parameter} heuristically obtains a global magnitude threshold along with the re-allocation of the weights based on the non-zero weights present at every step. Sparse Networks From Scratch (SNFS)~\citep{dettmers2019sparse} utilizes momentum of the weights to re-allocate weights across layers and the Rigged Lottery (RigL)~\citep{evci2019rigging} uses the magnitude to drop and the periodic dense gradients to regrow weights. SNFS and RigL are state-of-the-art in sparse-to-sparse training but fall short of GMP for the same experimental settings. It should be noted that, even though sparse-to-sparse can reduce the training cost, the existing frameworks~\citep{paszke2019pytorch, abadi2016tensorflow} consider the models as dense resulting in minimal gains.

DNW~\citep{wortsman2019discovering} and Dynamic Pruning with Feedback (DPF)~\citep{lin2020dynamic} fall between both as DNW uses a fully dense gradient in the backward pass and DPF maintains a copy of the dense model in parallel to optimize the sparse model through feedback. Note that DPF is complementary to most of the techniques discussed here.

\subsection{Uniform and Non-uniform Sparsity}
Uniform sparsity implies that all the layers in the DNN have the same amount of sparsity in proportion. Quite a few works have used uniform sparsity~\citep{gale2019state}, given its ease and lack of hyperparameters. However, some works keep parts of the model dense, including the first or the last layers~\citep{lin2020dynamic, mostafa2019parameter, zhu2017prune}. In general, making the first or the last layers dense benefits all the methods. GMP typically uses uniform sparsity and achieves state-of-the-art results.

Non-uniform sparsity permits different layers to have different sparsity budgets. Weight re-allocation heuristics have been used for non-uniform sparsity in DSR and SNFS. It can be a fixed budget like the ERK (Erdos-Renyi-Kernel) heuristic described in RigL~\citep{evci2019rigging}. A global pruning threshold~\citep{han2015learning} can also induce non-uniform sparsity and has been leveraged in Iterative Magnitude Pruning (IMP)~\citep{frankle2018the, renda2020comparing}. A good non-uniform sparsity budget can help in maintaining accuracy while also reducing the FLOPs due to a better parameter distribution. The aforementioned methods with non-uniform sparsity do not reduce the FLOPs compared to uniform sparsity in practice. Very few techniques like AMC~\citep{he2018amc}, using expensive reinforcement learning, minimize FLOPs with non-uniform sparsity. 

Most of the discussed techniques rely on intelligent heuristics to obtain non-uniform sparsity. Learning the pruning thresholds and in-turn learning the non-uniform sparsity budget is the main contribution of this paper.

\subsection{Learnable Sparsity}
Concurrent to our work,~\citep{savarese2019winning, Liu2020Dynamic, lee2019differentiable, xiao2019autoprune, azarian2020learned} have proposed learnable sparsity methods through training of the sparse masks and weights simultaneously with minimal heuristics. The reader is urged to review these works for a more complete picture of the field. Note that, while \alg is proposed to induce layer-wise unstructured sparsity, it can be easily adapted for global, filter-wise, or per-weight sparsity as discussed in Appendix~\ref{sec:adapt}. 
\section{Method - \alg}
\label{sec:method}
Optimization under sparsity constraint on the parameter set is a well studied area spanning more than three decades \cite{donoho1995noising,candes2007dantzig,jain2014iterative}, and is modeled as:$$\min_{\Wcal} \Lcal(\Wcal;\Dcal),\ \text{s.t.}\ \|\Wcal\|_0\leq k,$$ where $\Dcal \coloneqq \curly{\x_i\in\Rd{d},y_i\in\R}_{i\in\squarebrack{n}}$ is the observed data, $\Lcal$ is the loss function, $\Wcal$ are the parameters to be learned and $\|\cdot\|_0$ denotes the $L_0$-norm or the number of non-zeros, and $k$ is the parameter budget. Due to non-convexity and combinatorial structure of the $L_0$ norm constraint, it's convex relaxation $L_1$ norm has been studied for long time and has been at the center of a large literature on high-dimensional learning. In particular, several methods have been proposed to solve the two problems including projected gradient descent, forward/backward pruning etc. 

Projected Gradient Descent (PGD) in particular has been popular for both the problems as the projection onto both $L_0$ as well as the $L_1$ ball is computable in almost closed form \cite{beck2009fast,jain2014iterative}; $L_0$ ball projection is called Hard Thresholding while $L_1$ ball projection is known as Soft Thresholding. Further, these methods have been the guiding principle for many modern DNN model pruning (sparsity) techniques~\citep{han2015learning,zhu2017prune,narang2017exploring}. 

However, projection-based methods suffer from the problem of dense gradient and intermediate parameter structure, as the gradient descent iterate can be arbitrarily out of the set and is then projected back onto $L_0$ or $L_1$ ball. At a scale of billions of parameters, computing such dense gradients and updates can be daunting. More critically, the budget parameter $k$ is set at the global level, so it is not clear how to partition the budget for each layer, as the importance of each layer can be significantly different. 

In this work, we propose a reparameterization, Soft Threshold Reparameterization (\alg) based on the soft threshold operator~\citep{donoho1995noising}, to alleviate both the above mentioned concerns. That is, instead of first updating $\Wcal$ via gradient descent and then computing its projection, we directly optimize over projected $\Wcal$. Let $\Scal_g(\Wcal;s)$ be the projection of $\Wcal$ parameterized by $s$ and function $g$. $\Scal$ is applied to each element of $\Wcal$ and is defined as:
\begin{equation}
\Scal_g(w,s) := \sign{w}\cdot\mathrm{ReLU}(\abs{w}-g(s)),\label{eq:Sg}
\end{equation}
where $s$ is a learnable parameter, $g:\mathbb{R}\rightarrow \mathbb{R}$, and $\alpha = g(s)$ is the pruning threshold. $\mathrm{ReLU}(a)=\max(a,0)$. That is, if $|w|\leq g(s)$, then $\Scal_g(w,s)$ sets it to $0$. 

Reparameterizing the optimization problem with $\Scal$ {\em modifies} (note that it is not equivalent) it to: 
\begin{equation}\label{eq:str_a} \min_{\Wcal} \Lcal(\Scal_g(\Wcal,\bm{s}),\Dcal).\end{equation}

For $L$-layer DNN architectures, we divide $\Wcal$ into: $\Wcal = \squarebrack{\W_l}_{l=1}^L$ where $\W_l$ is the parameter tensor for the $l$-th layer. 
As mentioned earlier, different layers of DNNs are unique can have significantly different number of parameters. Similarly, different layers might need different sparsity budget for the best accuracy.  So, we set the trainable pruning parameter for each layer as $s_l$. That is, $\bm{s}=[s_1, \dots, s_L]$. 

Now, using the above mentioned reparameterization for each $\W_l$ and adding a standard $L_2$ regularization per layer, we get the following Gradient Descent (GD) update equation at the $t$-th step for $\W_l,\ \forall\ l\in \squarebrack{L}$:
\begin{align}
&\W_{l}^{(t+1)} \gets (1-\eta_t\cdot \lambda)\W_{l}^{(t)} \nonumber\\&\hspace*{-8pt}- \eta_t\nabla_{\Scal_g(\W_{l},s_l) }\Lcal(\Scal_g(\Wcal^{(t)},\bm{s}),\Dcal) \odot \nabla_{\W_l}\Scal_g(\W_l,s_l),\ \label{eq:gd}
\end{align}
where $\eta_t$ is the learning rate at the $t$-th step, and $\lambda$ is the $L_2$ regularization (weight-decay) hyper-parameter. $\nabla_{\W_{l} }\Scal_g(\W_l,s_l)$ is the gradient of $\Scal_g(\W_l,s_l)$ w.r.t. $\W_l$.

Now, $\Scal$ is non-differentiable, so we use sub-gradient which leads to the following update equation: {\small
\begin{align}
&\W_{l}^{(t+1)} \gets (1-\eta_t\cdot \lambda)\W_{l}^{(t)} \nonumber\\
&- \eta_t\nabla_{\Scal_g(\W_{l},s_l) }\Lcal(\Scal_g(\Wcal^{(t)},\bm{s}),\Dcal)\odot\indicator{\Scal_g(\W^{(t)}_l,s_l)\neq 0},\label{eq:new_gradient}
\end{align}}
where $\indicator{\cdot}$ is the indicator function and $A\odot B$ denotes element-wise (Hadamard) product of tensors $A$ and $B$. 

Now, if $g$ is a continuous function, then using the \alg \eqref{eq:str_a} and \eqref{eq:Sg}, it is clear that $\Lcal(\Scal_g(\Wcal,\bm{s}),\Dcal)$ is a continuous function of $\bm{s}$. Further, sub-gradient of $\Lcal$ w.r.t. $\bm{s}$, can be computed and uses for gradient descent on $\bm{s}$ as well; see Appendix~\ref{sec:sl_update}. Algorithm~\ref{alg:code} in the Appendix shows the implementation of \alg on 2D convolution along with extensions to global, per-filter \& per-weight sparsity. \alg can be modified and applied on the eigenvalues of a parameter tensor, instead of individual entries mentioned above, resulting in low-rank tensors; see Section~\ref{sec:RNN} for further details. Note that $\bm{s}$ also has the same weight-decay parameter $\lambda$.

\begin{figure}[!ht]
	\includegraphics[width=\columnwidth]{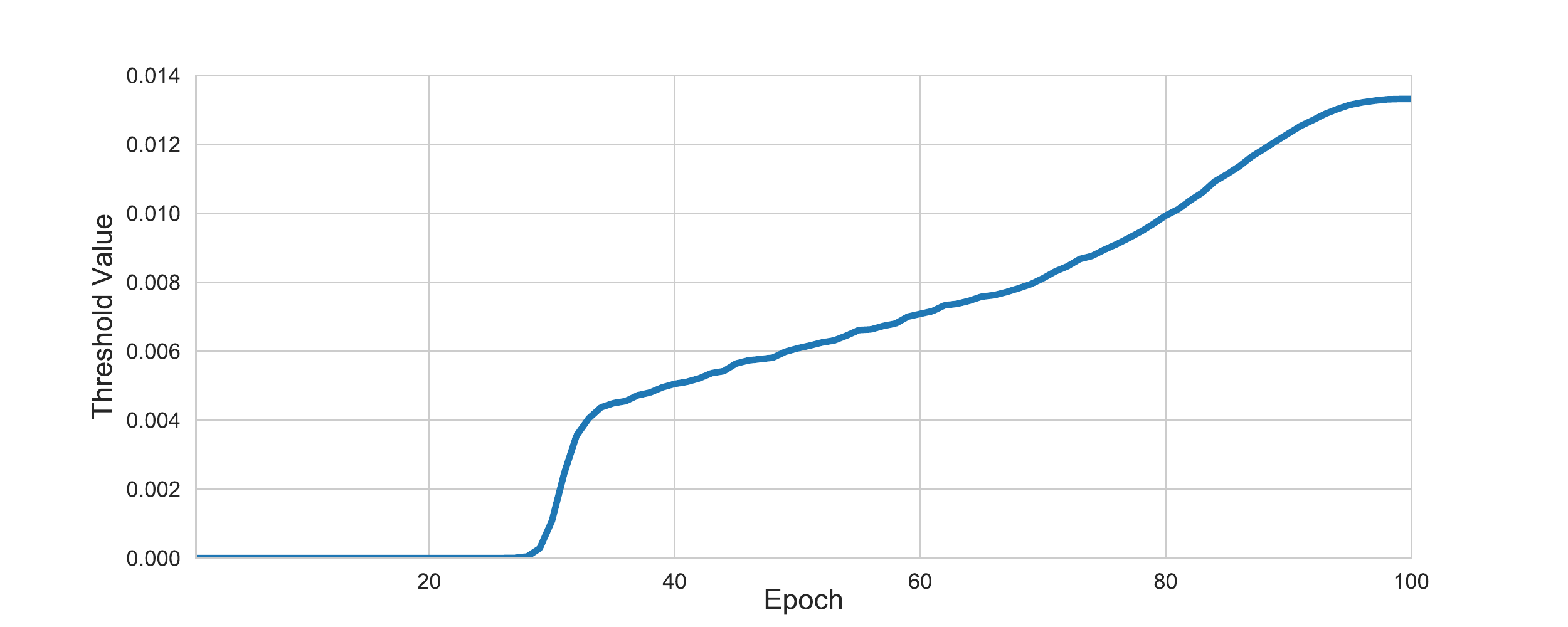}\vspace{-4mm}
	\caption{The learnt threshold parameter, $\alpha = g(s)$, for layer 10 in 90\% sparse ResNet50 on ImageNet-1K over the course of training.}
	\label{fig:tvse}
\end{figure}

\begin{figure}[!ht]
	\includegraphics[width=\columnwidth]{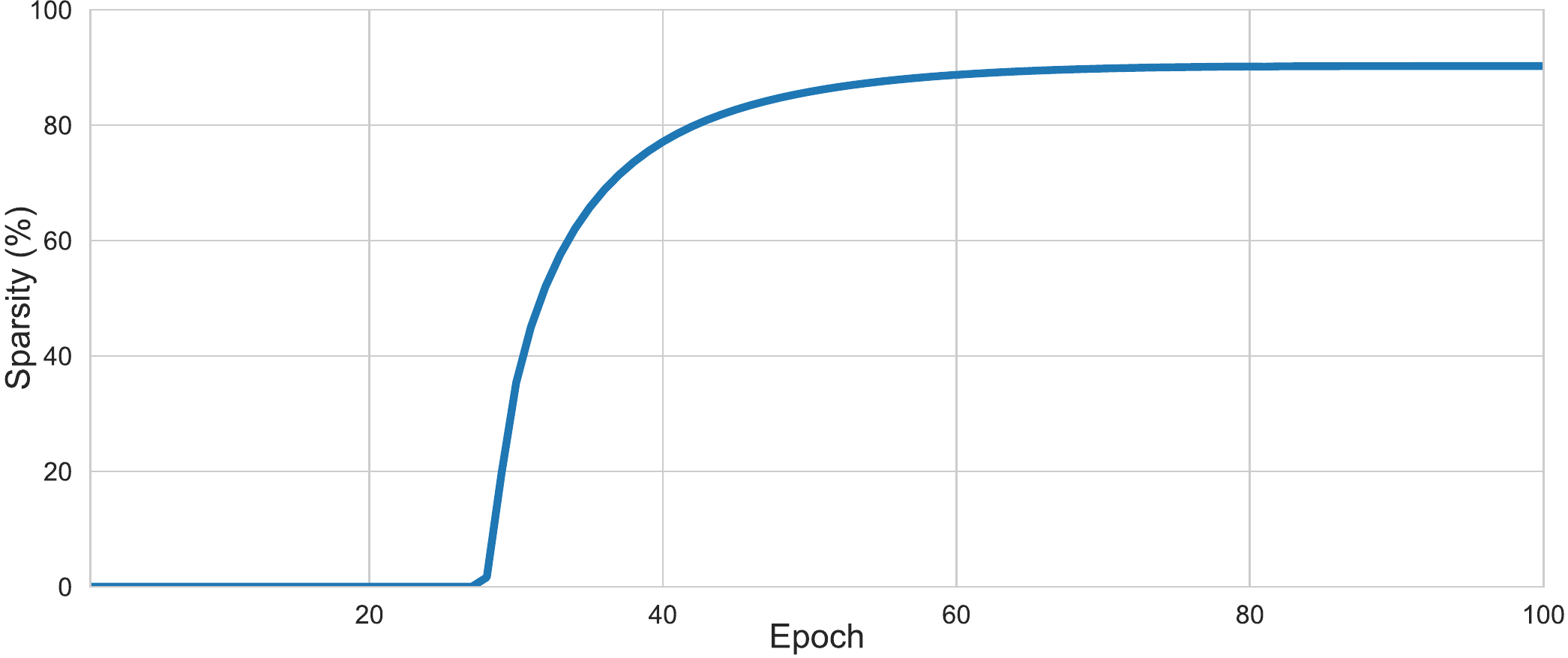}\vspace{-4mm}
	\caption{The progression of the learnt overall budget for 90\% sparse ResNet50 on ImageNet-1K over the course of training.}
 	\label{fig:svse}
\end{figure}

Naturally, $g$ plays a critical role here, as a sharp $g$ can lead to an arbitrary increase in threshold leading to poor accuracy while a flat $g$ can lead to slow learning. Practical considerations for choice of $g$ are discussed in Appendix~\ref{sec:g}. For the experiments, $g$ is set as the Sigmoid function for unstructured sparsity and the exponential function for structured sparsity.  Typically, $\curly{s_l}_{l\in\squarebrack{L}}$ are initialized with $s_{\rm init}$ to ensure that the thresholds $\curly{\alpha_l = g(s_l)}_{l\in\squarebrack{L}}$ start close to $0$. Figure~\ref{fig:tvse} shows that the thresholds' dynamics are guided by a combination of gradients from $\Lcal$ and the weight-decay on $\bm{s}$. Further, the overall sparsity budget for \alg is not set explicitly. Instead, it is controlled by the weight-decay parameter ($\lambda$), and can be further fine-tuned using $s_{\rm init}$. Interestingly, this curve is similar to the handcrafted heuristic for thresholds defined in~\citep{narang2017exploring}. Figure~\ref{fig:svse} shows the overall learnt sparsity budget for ResNet50 during training. The curve looks similar to GMP~\citep{zhu2017prune} sparsification heuristic, however, \alg learns it via backpropagation and SGD.
\begin{figure}[!ht]
	\includegraphics[width=\columnwidth]{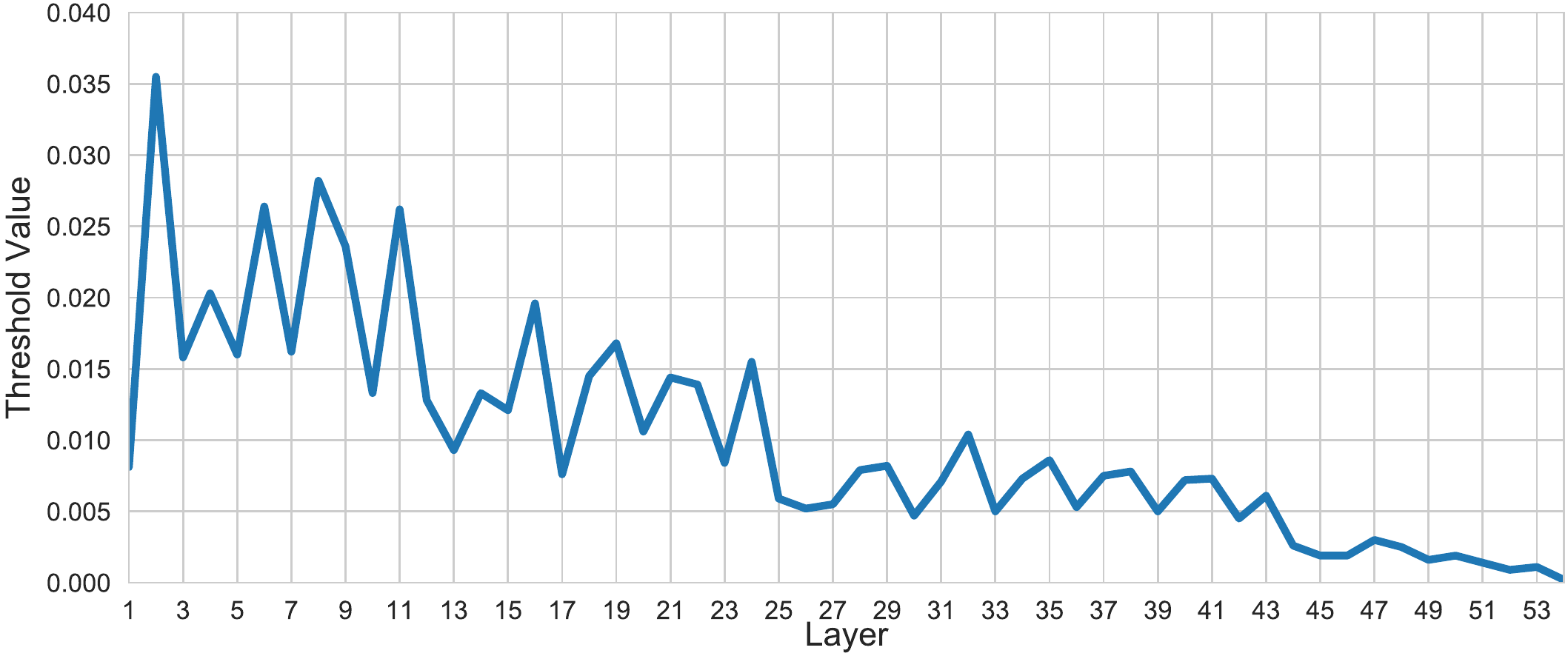}\vspace{-4mm}
	\caption{The final learnt threshold values, $[\alpha_l]^{54}_{l=1} = [g(s_l)]^{54}_{l=1}$, for all the layers in 90\% sparse ResNet50 on ImageNet-1K.}
 	\label{fig:tvsl}
\end{figure}

Finally, each parameter tensor learns a different threshold value, $\curly{\alpha_l}_{l\in\squarebrack{L}}$, resulting in unique final thresholds across the layers, as shown in Figure~\ref{fig:tvsl} for ResNet50. This, in turn, results in the non-uniform sparsity budget (see Figure~\ref{fig:sd}) which is empirically shown to be effective in increasing prediction accuracy while reducing FLOPs. Moreover, \eqref{eq:new_gradient} shows that the gradient update itself is sparse as gradient of $\Lcal$ is multiplied with an  indicator function of $\Scal_g(\W_l)\neq 0$ which gets sparser over iterations (Figure~\ref{fig:svse}). So \alg addresses both the issues with standard PGD methods (Hard/Soft Thresholding) that we mentioned above. 

\subsection{Analysis}
\label{sec:analysis}
The reparameterization trick using the projection operator's functional form can be used for standard constrained optimization problems as well (assuming the projection operator has a closed-form). However, it is easy to show that in general, such a method need not converge to the optimal solution even for convex functions over convex sets. This raises a natural question about the effectiveness of the technique for sparse weights learning problem. It turns out that for sparsity constrained problems, \alg is very similar to backward pruning \cite{hastie2009elements} which is a well-known technique for sparse regression. Note that, similar to Hard/Soft Thresholding, standard backward pruning also does not support differentiable tuning thresholds which makes it challenging to apply it to DNNs.

To further establish this connection, let's consider a standard sparse regression problem where $\mathbf{y}=\mathbf{Xw}^*$, $\mathbf{X}_{ij}\sim \Ncal(0,1)$, and $\mathbf{X}\in \mathbb{R}^{n\times d}$. $\mathbf{w}^*\in\{0,1\}^d$ has $r\ll d$ non-zeros, and $d\gg n\gg r\log d$. Due to the initialization, $g(s)\approx 0$ in initial few iterations. So, gradient descent converges to the least $\ell_2$-norm regression solution. That is, $\mathbf{w}=\mathbf{UU}^T \mathbf{w}^*$ where $\mathbf{U}\in \mathbb{R}^{d\times n}$ is the right singular vector matrix of $\mathbf{X}$ and is a random $n$-dimensional subspace. As $\mathbf{U}$ is a random subspace. Since $n\gg r \log d$, $\mathbf{U}_{S}\mathbf{U}_S^T \approx \frac{r}{d}\cdot I$ where $S=\mathrm{supp}(\w^*)$, and $\mathbf{U}_S$ indexes rows of $\mathbf{U}$ corresponding to $S$. That is, $\min_{j\in S} \abs{\mathbf{U}_j\cdot \mathbf{U}^T \mathbf{w}^*}\geq 1-o(1)$. On the other hand, $\abs{\mathbf{U}_j\cdot \mathbf{U}_S^T \mathbf{w}^*} \lesssim \frac{\sqrt{nr}}{d}\sqrt{\log d}$ with high probability for $j\not\in S$. As $n\gg r \log d$, almost all the elements of $\mathrm{supp}(\w^*)$ will be in top $\BigO{n}$ elements of $\mathbf{w}$. Furthermore, $\mathbf{X}\Scal_g(\mathbf{w},s)=\y$, so $\abs{s}$ would decrease significantly via weight-decay and hence $g(s)$ becomes large enough to prune all but say $\BigO{n}$ elements. Using a similar argument as above, leads to further pruning of $\mathbf{w}$, while ensuring recovery of almost all elements in $\mathrm{supp}(\mathbf{w}^*)$.

\section{Experiments}
\label{sec:expts}
This section showcases the experimentation followed by the observations from applying \alg for (a) unstructured sparsity in CNNs and (b) structured sparsity in RNNs.

\subsection{Unstructured Sparsity in CNNs}
\subsubsection{Experimental Setup}
ImageNet-1K~\citep{deng2009imagenet} is a widely used large-scale image classification dataset with 1K classes. All the CNN experiments presented are on ImageNet-1K. ResNet50~\citep{he2016deep} and MobileNetV1~\citep{howard2017mobilenets} are two popular CNN architectures. ResNet50 is extensively used in literature to show the effectiveness of sparsity in CNNs. Experiments on MobileNetV1 argue for the generalizability of the proposed technique (\alg). Dataset and models' details can be found in Appendix~\ref{sec:datasetdet}. 

\alg was compared against strong state-of-the-art baselines in various sparsity regimes including GMP~\citep{gale2019state}, DSR~\citep{mostafa2019parameter}, DNW~\citep{wortsman2019discovering}, SNFS~\citep{dettmers2019sparse}, RigL~\citep{evci2019rigging} and DPF~\citep{lin2020dynamic}. GMP and DNW always use a uniform sparsity budget. RigL, SNFS, DSR, and DPF were compared in their original form. Exceptions for the uniform sparsity are marked in Table~\ref{tab:res50main}. The ``+ ERK" suffix implies the usage of ERK budget~\citep{evci2019rigging} instead of the original sparsity budget. Even though VD~\citep{molchanov2017variational} achieves state-of-the-art results, it is omitted due to the 2$\times$ memory and 4$\times$ compute footprint during training. Typically VD and IMP use a global threshold for global sparsity (GS)~\citep{han2015learning} which can also be learnt using \alg. The unstructured sparsity experiments presented compare the techniques which induce layer-wise sparsity. Note that \alg is generalizable to other scenarios as well. Open-source implementations, pre-trained models, and reported numbers of the available techniques were used as the baselines. Experiments were run on a machine with 4 NVIDIA Titan X (Pascal) GPUs.

All baselines use the hyperparameter settings defined in their implementations/papers. The experiments for \alg use a batch size of 256, cosine learning rate routine and are trained for 100 epochs following the hyperparameter settings in~\citep{wortsman2019discovering} using SGD + momentum. \alg has weight-decay ($\lambda$) and $s_{\rm init}$ hyperparameters to control the overall sparsity in CNNs and can be found in Appendix~\ref{sec:hyperparams}. $\text{GMP}_{1.5\times}$~\citep{gale2019state} and $\text{RigL}_{5\times}$~\citep{evci2019rigging} show that training the networks longer increases accuracy. However, due to the limited compute and environmental concerns~\citep{schwartz2019green}, all the experiments were run only for around 100 epochs ($\sim$3 days each). Unstructured sparsity in CNNs with \alg is enforced by learning one threshold per-layer as shown in Figure~\ref{fig:tvsl}. PyTorch \ensuremath{{\rm STRConv}} code can be found in Algorithm~\ref{alg:code} of Appendix.

\subsubsection{ResNet50 on ImageNet-1K}
\begin{table}[!b]
\centering\vspace{-6mm}
\caption{\alg is the state-of-the-art for unstructured sparsity in ResNet50 on ImageNet-1K while having lesser inference cost (FLOPs) than the baselines across all the sparsity regimes. $^*$ and $^\#$ imply that the first and last layer are dense respectively. Baseline numbers reported from their respective papers/open-source implementations and models. FLOPs do not include batch-norm.}
\label{tab:res50main}
\resizebox{\columnwidth}{!}{
\begin{tabular}{@{}lcrcr@{}}
\toprule
Method              & \begin{tabular}[c]{@{}c@{}}Top-1 Acc \\ (\%)\end{tabular} & Params & \begin{tabular}[c]{@{}c@{}}Sparsity \\ (\%)\end{tabular} & FLOPs      \\ \midrule
ResNet-50           & 77.01                                                     & 25.6M  & 0.00                                                     & 4.09G         \\\midrule
GMP                 & 75.60                                                     & 5.12M  & 80.00                                                    & 818M       \\
DSR$^{*\#}$                 & 71.60                                                     & 5.12M  & 80.00                                                    & 1.23G      \\
DNW                 & 76.00                                                     & 5.12M  & 80.00                                                    & 818M       \\
SNFS                & 74.90                                                     & 5.12M  & 80.00                                                    & -          \\
SNFS + ERK          & 75.20                                                     & 5.12M  & 80.00                                                    & 1.68G      \\
RigL$^{*}$                & 74.60                                                     & 5.12M  & 80.00                                                    & 920M       \\
RigL + ERK          & 75.10                                                     & 5.12M  & 80.00                                                    & 1.68G      \\
DPF                 & 75.13                                                     & 5.12M  & 80.00                                                    & 818M       \\
\textit{\alg} & \textbf{76.19}                                            &  5.22M      &    79.55                                                      & \textbf{766M}           \\
\textit{\alg} & \textbf{76.12}                                            &  \textbf{4.47M}      &    \textbf{81.27 }                                                     & \textbf{705M}           \\ \midrule
GMP                 & 73.91                                                     & 2.56M  & 90.00                                                    & 409M       \\
DNW                 & 74.00                                                     & 2.56M  & 90.00                                                    & 409M       \\
SNFS                & 72.90                                                     & 2.56M  & 90.00                                                    & 1.63G      \\
SNFS + ERK          & 72.90                                                     & 2.56M  & 90.00                                                    & 960M       \\
RigL$^{*}$                & 72.00                                                     & 2.56M  & 90.00                                                    & 515M       \\
RigL + ERK          & 73.00                                                     & 2.56M  & 90.00                                                    & 960M       \\
DPF$^{\#}$                 & 74.55                                                     & 4.45M  & 82.60                                                    & 411M       \\
\textit{\alg} & \textbf{74.73}                                                     & 3.14M  & 87.70                                                    & \textbf{402M}       \\
\textit{\alg} & \textbf{74.31}                                                     & \textbf{2.49M}  & \textbf{90.23}                                                    & \textbf{343M}       \\
\textit{\alg} & \textbf{74.01}                                                     & \textbf{2.41M}  & \textbf{90.55}                                                    & \textbf{341M}       \\ \midrule
GMP                 & 70.59                                                     & 1.28M  & 95.00                                                    & 204M       \\
DNW                 & 68.30                                                     & 1.28M  & 95.00                                                    & 204M       \\
RigL$^{*}$                & 67.50                                                      & 1.28M  & 95.00                                                    & 317M       \\
RigL + ERK          & 70.00                                                     & 1.28M  & 95.00                                                    & $\sim$600M \\
\textit{\alg} & \textbf{70.97}                                                     & 1.33M  & 94.80                                                    & \textbf{182M}       \\
\textit{\alg} & 70.40                                                      & \textbf{1.27M}  & \textbf{95.03}                                                    & \textbf{159M}       \\
\textit{\alg} & 70.23                                                     & \textbf{1.24M}  & \textbf{95.15}                                                    & \textbf{162M}       \\ \midrule
RigL$^{*}$                & 64.50                                                     & 0.90M  & 96.50                                                    & 257M       \\
RigL + ERK          & 67.20                                                     & 0.90M  & 96.50                                                    & $\sim$500M \\
\textit{\alg} & \textbf{67.78}                                                     & 0.99M  & 96.11                                                    & \textbf{127M}       \\
\textit{\alg} & \textbf{67.22}                                                     & \textbf{0.88M}  & \textbf{96.53}                                                    & \textbf{117M}       \\ \midrule
GMP                 & 57.90                                                     & 0.51M  & 98.00                                                    & 82M        \\
DNW                 & 58.20                                                     & 0.51M  & 98.00                                                    & 82M        \\
\textit{\alg} & \textbf{62.84}                                                     & 0.57M  & 97.78                                                    & \textbf{80M}        \\
\textit{\alg} & \textbf{61.46}                                                     & \textbf{0.50M}  & \textbf{98.05}                                                    & \textbf{73M}        \\
\textit{\alg} & \textbf{59.76}                                                     & \textbf{0.45M}  & \textbf{98.22}                                                    & \textbf{68M}        \\\midrule
GMP                 & 44.78                                                     & 0.26M  & 99.00                                                    & 41M        \\
\textit{\alg} & \textbf{54.79}                                                     & 0.31M  & 98.79                                                    & 54M        \\
\textit{\alg} & \textbf{51.82}                                                     & 0.26M  & 98.98                                                    & 47M        \\
\textit{\alg} & \textbf{50.35}                                                     & \textbf{0.23M}  & \textbf{99.10}                                                    & 44M        \\ \bottomrule

\end{tabular}}
\end{table}
A fully dense ResNet50 trained on ImageNet-1K has 77.01\% top-1 validation accuracy. \alg is compared extensively to other baselines on ResNet50 in the sparsity ranges of 80\%, 90\%, 95\%, 96.5\%, 98\%, and 99\%. Table~\ref{tab:res50main} shows that DNW and GMP are state-of-the-art among the baselines across all the aforementioned sparsity regimes. As \alg might not be able to get exactly to the sparsity budget, numbers are reported for the models which nearby. Note that the 90.23\% sparse ResNet50 on ImageNet-1K with \alg is referred to as the 90\% sparse ResNet50 model learnt with \alg.

\begin{figure}[!ht]
	\includegraphics[width=\columnwidth]{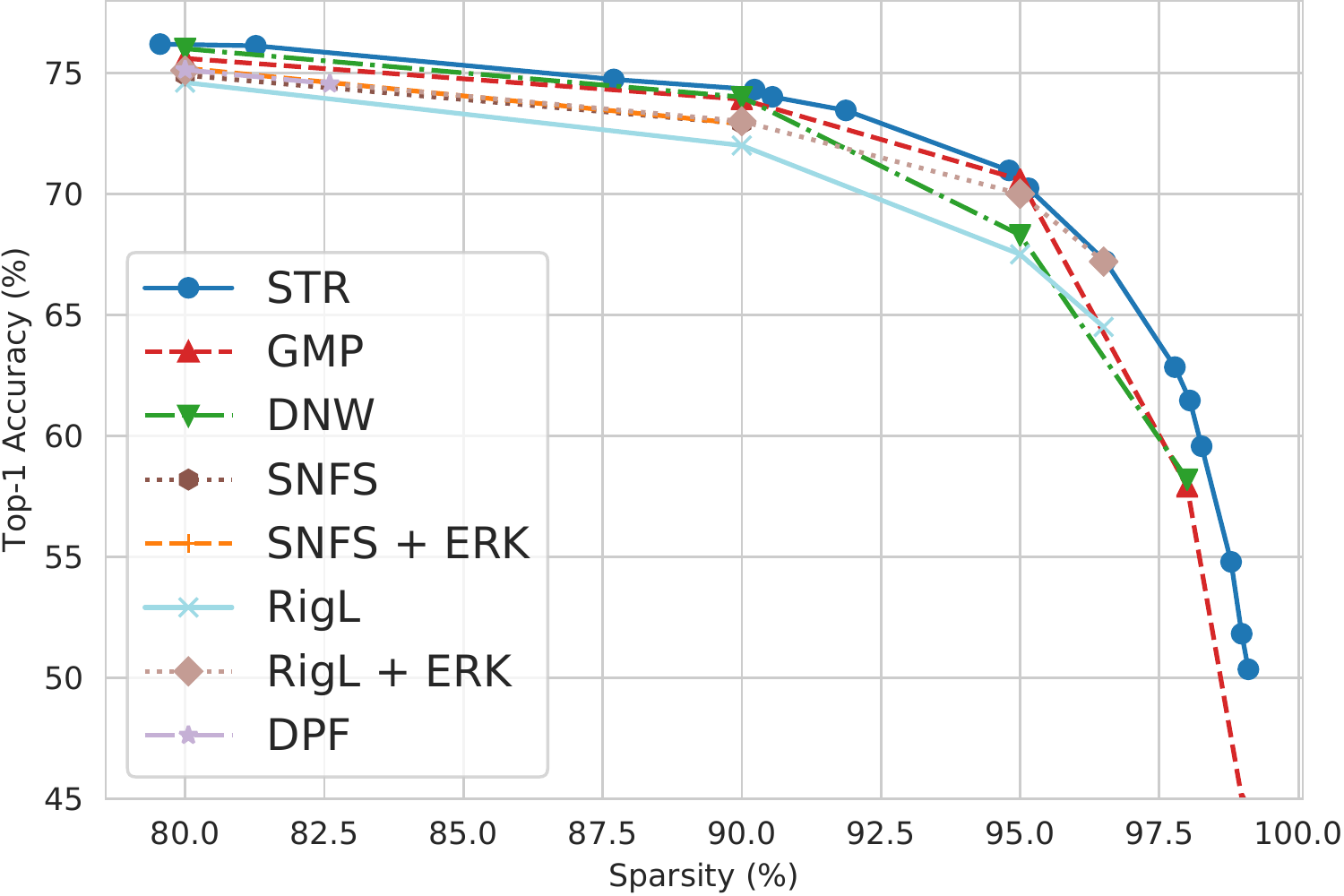}\vspace{-4mm}
	\caption{\alg forms a frontier curve over all the baselines in all sparsity regimes showing that it is the state-of-the-art for unstructured sparsity in ResNet50 on ImageNet-1K.}
 	\label{fig:avss}
\end{figure}
\begin{figure}[!ht]
	\includegraphics[width=\columnwidth]{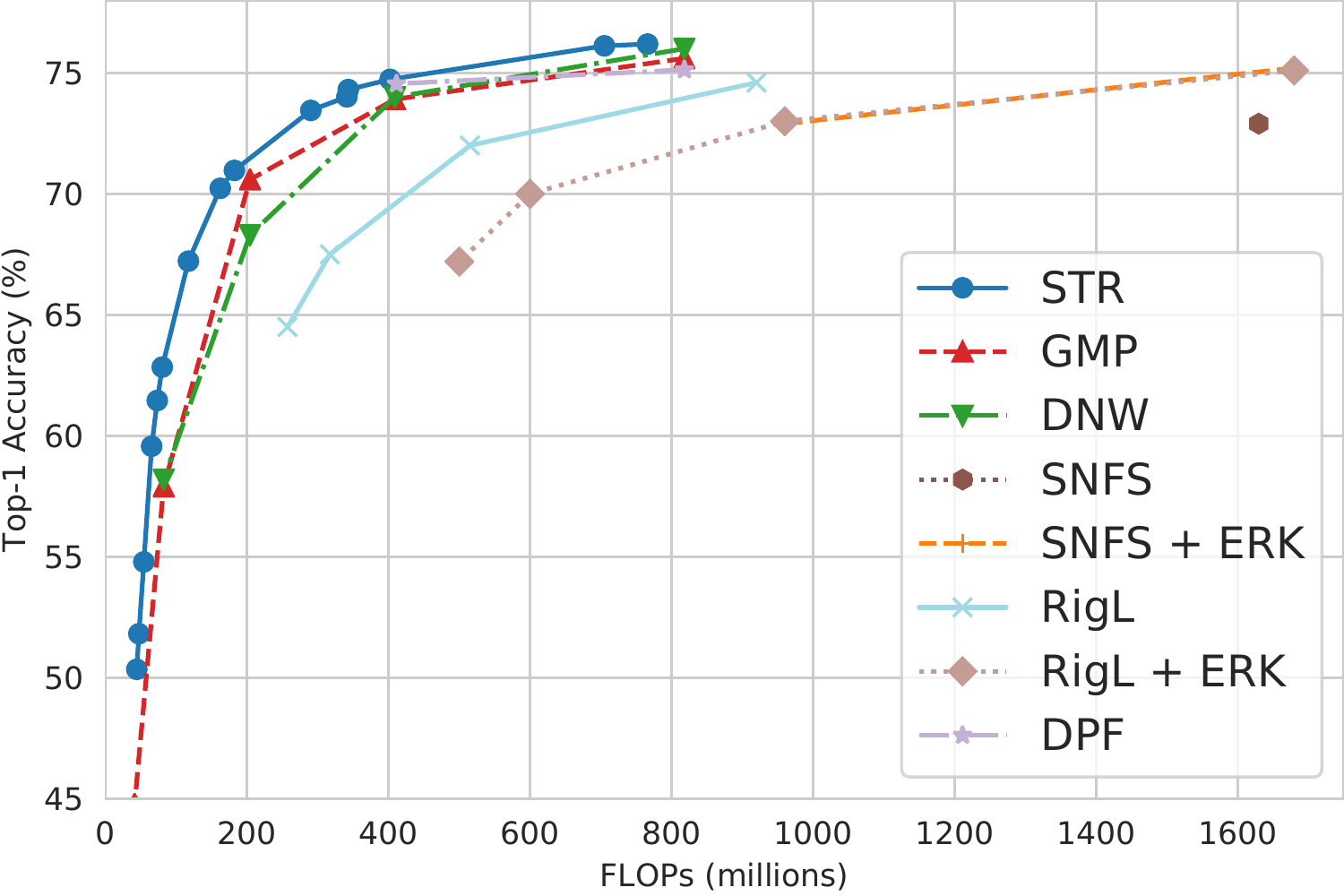}\vspace{-4mm}
	\caption{\alg results in ResNet50 models on ImageNet-1K which have the lowest inference cost (FLOPs) for any given accuracy.}
 	\label{fig:avsf}
\end{figure}

\alg comfortably beats all the baselines across all the sparsity regimes as seen in Table~\ref{tab:res50main} and is the state-of-the-art for unstructured sparsity. Figure~\ref{fig:avss} shows that \alg forms a frontier curve encompassing all the baselines at all the levels of sparsity. Very few methods are stable in the ultra sparse regime of 98-99\% sparsity and GMP can achieve 99\% sparsity. \alg is very stable even in the ultra sparse regime, as shown in Table~\ref{tab:res50main} and Figure~\ref{fig:avss}, while being up to 10\% higher in accuracy than GMP at 99\% sparsity.

\alg induces non-uniform sparsity across layers, Table~\ref{tab:res50main} and Figure~\ref{fig:avsf} show that \alg produces models which have lower or similar inference FLOPs compared to the baselines while having better prediction accuracy in all the sparsity regimes. This hints at the fact that \alg could be redistributing the parameters thereby reducing the FLOPs. In the 80\% sparse models, \alg is at least 0.19\% better in accuracy than the baselines while having at least 60M (6.5\%) lesser FLOPs. Similarly, \alg has state-of-the-art accuracy in 90\%, 95\%, and 96.5\% sparse regimes while having at least 68M (16.5\%), 45M (22\%) and 140M (54\%) lesser FLOPs than the best baselines respectively. In the ultra sparse regime of 98\% and 99\% sparsity, \alg has similar or slightly higher FLOPs compared to the baselines but is up to 4.6\% and 10\% better in accuracy respectively. Table~\ref{tab:res50main} summarizes that the non-uniform sparsity baselines like SNFS, SNFS+ERK, and RigL+ERK can have up to 2-4$\times$ higher inference cost (FLOPs) due to non-optimal layer-wise distribution of the parameter weights.

\begin{figure}[!ht]
	\includegraphics[width=\columnwidth]{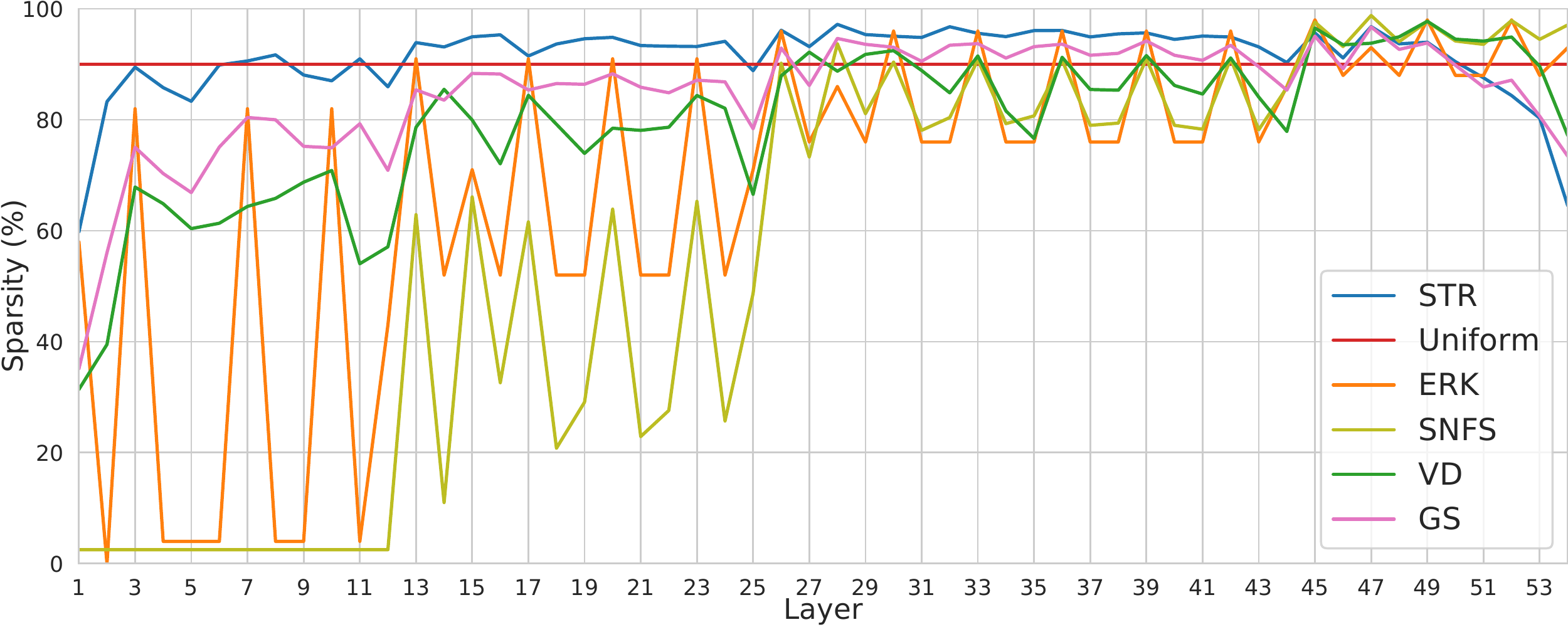}\vspace{-4mm}
	\caption{Layer-wise sparsity budget for the 90\% sparse ResNet50 models on ImageNet-1K using various sparsification techniques.}
 	\label{fig:sd}
\end{figure}
\begin{figure}[!ht]
	\includegraphics[width=\columnwidth]{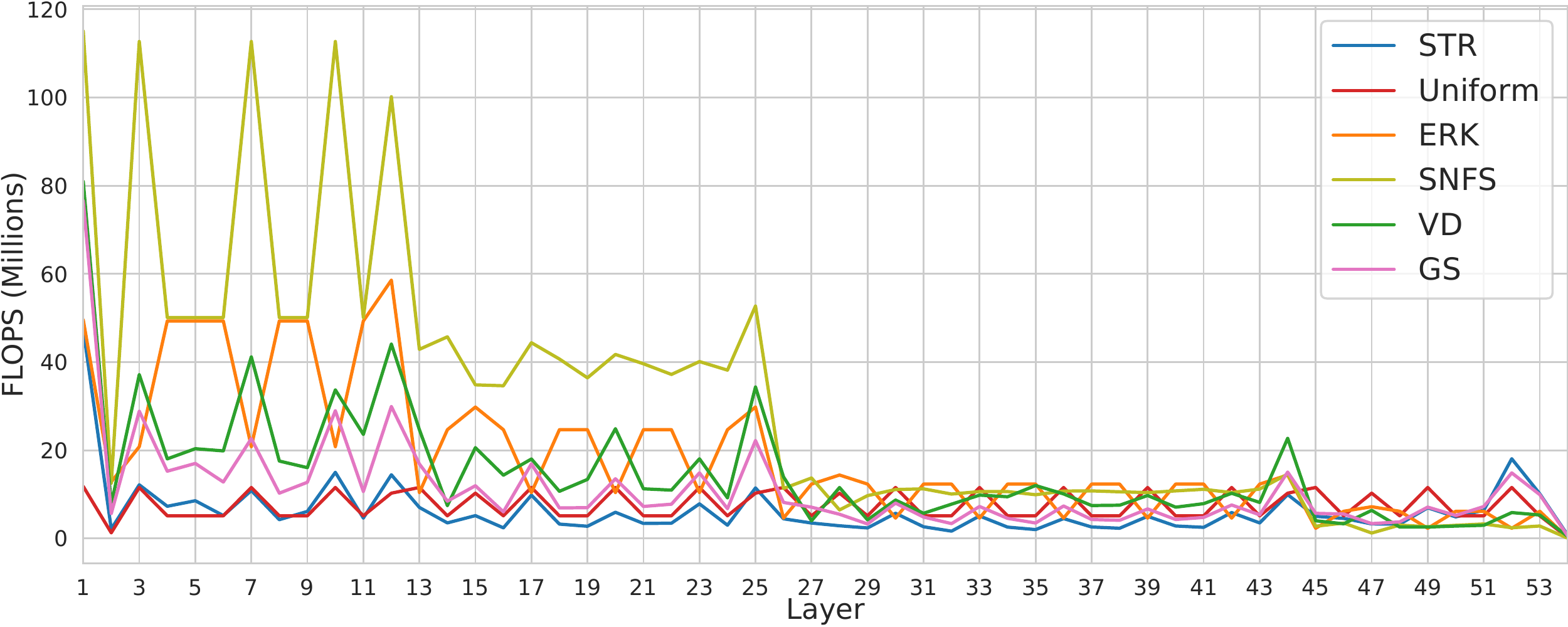}\vspace{-4mm}
	\caption{Layer-wise FLOPs budget for the 90\% sparse ResNet50 models on ImageNet-1K using various sparsification techniques.}
 	\label{fig:fd}
\end{figure}
\label{sec:observations}
\textbf{Observations:} \alg on ResNet50 shows some interesting observations related to sparsity and inference cost (FLOPs). These observations will be further discussed in Section~\ref{sec:disc}:\vspace{-3mm}
\begin{enumerate}[leftmargin=*]
    \itemsep 0pt
    \topsep 0pt
    \parskip 2pt
    \item \alg is state-of-the-art for unstructured sparsity.
    \item \alg minimizes inference cost (FLOPs) while maintaining accuracy in the 80-95\% sparse regime.
    \item \alg maximizes accuracy while maintaining inference cost (FLOPs) in 98-99\% ultra sparse regime.
    \item \alg learns a non-uniform layer-wise sparsity, shown in Figure~\ref{fig:sd}, which shows that the initial layers of the CNN can be sparser than that of the existing non-uniform sparsity methods. All the learnt non-uniform budgets through \alg can be found in Appendix~\ref{sec:res50budget}.
    \item Figure~\ref{fig:sd} also shows that the last layers through \alg are denser than that of the other methods which is contrary to the understanding in the literature of non-uniform sparsity~\citep{mostafa2019parameter, dettmers2019sparse, evci2019rigging, gale2019state}. This leads to a sparser backbone for transfer learning. The backbone sparsities can be found in Appendix~\ref{sec:res50budget}.
    \item Figure~\ref{fig:fd} shows the layer-wise FLOPs distribution for the non-uniform sparsity methods. \alg adjusts the FLOPs across layers such that it has lower FLOPs than the baselines. Note that the other non-uniform sparsity budgets lead to heavy compute overhead in the initial layers due to denser parameter tensors.
\end{enumerate}
\alg can also induce global sparsity (GS)~\citep{han2015learning} with similar accuracy at $\sim2\times$ FLOPs compared to layer-wise for 90-98\% sparsity (details in Appendix~\ref{sec:global}).

\subsubsection{MobileNetV1 on ImageNet-1K}

MobileNetV1 was trained on ImageNet-1K for unstructured sparsity with \alg to ensure generalizability. Since GMP is the state-of-the-art baseline as shown earlier, \alg was only compared to GMP for 75\% and 90\% sparsity regimes. A fully dense MobileNetV1 has a top-1 accuracy of 71.95\% on ImageNet-1K. GMP~\citep{zhu2017prune} has the first layer and depthwise convolution layers dense for MobileNetV1 to ensure training stability and maximize accuracy.
\begin{table}[t]
\centering
\caption{\alg is up to 3\% higher in accuracy while having 33\% lesser inference cost (FLOPs) for MobileNetV1 on ImageNet-1K.}
\label{tab:mobilenetv1res}
\resizebox{\columnwidth}{!}{
\begin{tabular}{@{}lcccr@{}}
\toprule
Method              & \begin{tabular}[c]{@{}c@{}}Top-1 Acc \\ (\%)\end{tabular} & Params & \begin{tabular}[c]{@{}c@{}}Sparsity \\ (\%)\end{tabular} & FLOPs \\ \midrule
MobileNetV1         & 71.95                                                     & 4.21M  & 0.00                                                     & 569M  \\ \midrule
GMP                 & 67.70                                                     & 1.09M  & 74.11                                                    & 163M  \\
\textit{\alg} & \textbf{68.35}                                                     & \textbf{1.04M}  & \textbf{75.28}                                                     & \textbf{101M}  \\
\textit{\alg} & 66.52                                                     & \textbf{0.88M}  & \textbf{79.07}                                                    &\textbf{81M}   \\ \midrule
GMP                 & 61.80                                                     & 0.46M  & 89.03                                                    & 82M   \\
\textit{\alg} & \textbf{64.83}                                                     & 0.60M  & 85.80                                                    & \textbf{55M}   \\
\textit{\alg} & \textbf{62.10}                                                     & 0.46M  & 89.01                                                    & \textbf{42M}   \\
\textit{\alg} & 61.51                                                     & \textbf{0.44M}  & \textbf{89.62}                                                    & \textbf{40M}   \\ \bottomrule
\end{tabular}}
\end{table}

Table~\ref{tab:mobilenetv1res} shows the \alg is at least 0.65\% better than GMP for 75\% sparsity, while having at least 62M (38\%) lesser FLOPs. More interestingly, \alg has state-of-the-art accuracy while having up to 50\% (40M) lesser FLOPs than GMP in the 90\% sparsity regime. All the observations made for ResNet50 hold for MobileNetV1 as well. The sparsity and FLOPs distribution across layers can be found in Appendix~\ref{sec:mbv1dist}.

\subsection{Structured Sparsity in RNNs}
\subsubsection{Experimental Setup}
\label{sec:RNN}
Google-12 is a speech recognition dataset that has 12 classes made from the Google Speech Commands dataset~\citep{warden2017}. HAR-2 is a binarized version of the 6-class Human Activity Recognition dataset~\citep{anguita2012}. These two datasets stand as compelling cases for on-device resource-efficient machine learning at the edge. Details about the datasets can be found in Appendix~\ref{sec:datasetdet}.

FastGRNN~\citep{kusupati2018fastgrnn} was proposed to enable powerful RNN models on resource-constrained devices. FastGRNN relies on making the RNN parameter matrices low-rank, sparse and quantized. As low-rank is a form of structured sparsity, experiments were done to show the effectiveness of \alg for structured sparsity. The input vector to the RNN at each timestep and hidden state have $D$ \& $\hat{D}$ dimensionality respectively. FastGRNN has two parameter matrices, $\mathbf{W}\in\R^{D\times\hat{D}}$, $\mathbf{U}\in\R^{\hat{D}\times\hat{D}}$ which are reparameterized as product of low-rank matrices, $\mathbf{W} = \mathbf{W}_{1}\mathbf{W}_{2}$, and $\mathbf{U} = \mathbf{U}_{1}\mathbf{U}_{2}$ where  $\mathbf{W}_{1}\in\R^{D\times {r_W}}$,  $\mathbf{W}_{2}\in\R^{{r_W}\times\hat{D}}$, and  $(\mathbf{U}_{1})^\top, \mathbf{U}_{2}\in\R^{{r_U}\times\hat{D}}$. $r_W$, $r_U$ are the ranks of the respective matrices. In order to apply \alg, the low-rank reparameterization can be changed to $\mathbf{W} = (\mathbf{W}_{1}\odot\one\mathbf{m}_{\mathbf{W}}^\top )\mathbf{W}_{2}$, and $\mathbf{U} = (\mathbf{U}_{1}\odot\one\mathbf{m}_{\mathbf{U}}^\top)\mathbf{U}_{2}$ where $\mathbf{m}_{\mathbf{W}} = \one_{D}$, and $\mathbf{m}_{\mathbf{U}} = \one_{\hat{D}}$, $\mathbf{W}_{1}\in\R^{D\times D}$,  $\mathbf{W}_{2}\in\R^{D\times\hat{D}}$, and  $\mathbf{U}_{1}, \mathbf{U}_{2}\in\R^{\hat{D}\times\hat{D}}$. To learn the low-rank, \alg is applied on the $\mathbf{m}_{\mathbf{W}}$, and $\mathbf{m}_{\mathbf{U}}$ vectors. Learning low-rank with \alg on $\mathbf{m}_{\mathbf{W}}$, $\mathbf{m}_{\mathbf{U}}$ can be thought as inducing unstructured sparsity on the two trainable vectors aiming for the right $r_W$, and $r_U$.

The baseline is low-rank FastGRNN where the ranks of the matrices are preset~\citep{kusupati2018fastgrnn}. EdgeML~\citep{edgeml03} FastGRNN was used for the experiments with the hyperparameters suggested in the paper and is referred to as vanilla training. Hyperparameters for the models can be found in Appendix~\ref{sec:hyperparams}.

\subsubsection{FastGRNN on Google-12 and HAR-2}
Table~\ref{tab:rnnresults} presents the results for low-rank FastGRNN with vanilla training and \alg. Full-rank non-reparameterized FastGRNN has an accuracy of 92.60\% and 96.10\% on Google-12 and HAR-2 respectively. 
\begin{table}[!ht]
\caption{\alg can induce learnt low-rank in FastGRNN resulting in up to 2.47\% higher accuracy than the vanilla training.}
\label{tab:rnnresults}
\resizebox{1\columnwidth}{!}{
\begin{tabular}{@{}ccc|ccc@{}}
\toprule
\multicolumn{3}{c}{Google-12}                                                                                                        & \multicolumn{3}{|c}{HAR-2}                                                                                                         \\ \midrule
\multicolumn{1}{c|}{\multirow{2}{*}{($r_W$, $r_U$)}}                 & \multicolumn{2}{c}{Accuracy (\%)}                                   & \multicolumn{1}{|c|}{\multirow{2}{*}{($r_W$, $r_U$)}}               & \multicolumn{2}{c}{Accuracy (\%)}                                   \\ \cmidrule(lr){2-3} \cmidrule(l){5-6} 
\multicolumn{1}{c|}{}                                          & \begin{tabular}[c]{@{}c@{}}Vanilla \\ Training\end{tabular} & \alg  & \multicolumn{1}{c|}{}                                        & \begin{tabular}[c]{@{}c@{}}Vanilla \\ Training\end{tabular} & \alg  \\ \midrule
\begin{tabular}[c]{@{}c@{}}Full rank (32, 100)\end{tabular} & 92.30                                                       & -     & \begin{tabular}[c]{@{}c@{}}Full rank (9, 80)\end{tabular} & 96.10                                                       & -     \\\midrule
(12, 40)                                                       & 92.79                                                       & \textbf{94.45} & (9, 8)                                                       & 94.06                                                       & \textbf{95.76} \\
(11, 35)                                                       & 92.86                                                       & \textbf{94.42} & (9, 7)                                                       & 93.15                                                       & \textbf{95.62} \\
(10, 31)                                                       & 92.86                                                       & \textbf{94.25} & (8, 7)                                                       & 94.88                                                       & \textbf{95.59} \\
(9, 24)                                                        & 93.18                                                       & \textbf{94.45} &                                                        &                                                        &      \\ \bottomrule
\end{tabular}}
\end{table}
\alg outperforms vanilla training by up to 1.67\% in four different model-size reducing rank settings on Google-12. Similarly, on HAR-2, \alg is better than vanilla training in all the rank settings by up to 2.47\%. Note that the accuracy of the low-rank models obtained by \alg is either better or on-par with the full rank models while being around 50\% and 70\% smaller in size (low-rank) for Google-12 and HAR-2 respectively.

These experiments for structured sparsity in RNNs show that \alg can be applied to obtain low-rank parameter tensors. Similarly, \alg can be extended for filter/channel pruning and block sparsity~\citep{he2017channel, huang2018data,liu2018rethinking} and details for this adaptation can be found in Appendix~\ref{sec:structured}.

\section{Discussion and Drawbacks}
\label{sec:disc}
\alg's usage for unstructured sparsity leads to interesting observations as noted in Section~\ref{sec:observations}. It is clear from Table~\ref{tab:res50main} and Figures~\ref{fig:avss},~\ref{fig:avsf} that \alg achieves state-of-the-art accuracy for all the sparsity regimes and also reduces the FLOPs in doing so. \alg helps in learning non-uniform sparsity budgets which are intriguing to study as an optimal non-uniform sparsity budget can ensure minimization of FLOPs while maintaining accuracy. Although it is not clear why \alg's learning dynamics result in a non-uniform budget that minimizes FLOPs, the reduction in FLOPs is due to the better redistribution of parameters across layers. 

Non-uniform sparsity budgets learnt by \alg have the initial and middle layers to be sparser than the other methods while making the last layers denser. Conventional wisdom suggests that the initial layers should be denser as the early loss of information would be hard to recover, this drives the existing non-uniform sparsity heuristics.  As most of the parameters are present in the deeper layers, the existing methods tend to make them sparser while not affecting the FLOPs by much. \alg, on the other hand, balances the FLOPs and sparsity across the layers as shown in Figures~\ref{fig:sd},~\ref{fig:fd} making it a lucrative and efficient choice. The denser final layers along with sparser initial and middle layers point to sparser CNN backbones obtained using \alg. These sparse backbones can be viable options for efficient representation/transfer learning for downstream tasks.
\begin{table}[!ht]
\caption{Effect of various layer-wise sparsity budgets when used with DNW for ResNet50 on ImageNet-1K.}
\label{tab:res50budgettrans}
\resizebox{\columnwidth}{!}{
\begin{tabular}{@{}lcccc@{}}
\toprule
Method                          & \multicolumn{1}{c}{\begin{tabular}[c]{@{}c@{}}Top-1 Acc \\ (\%)\end{tabular}} & Params & \begin{tabular}[c]{@{}c@{}}Sparsity \\ (\%)\end{tabular} & FLOPs \\ \midrule
Uniform                         & 74.00                                                                          & 2.56M  & 90.00                                                    & 409M  \\
ERK                             & \textbf{74.10}                                                                          & 2.56M  & 90.00                                                       & 960M  \\
Budget from \alg & 74.01                                                                          & \textbf{2.49M}  & \textbf{90.23}                                                    & \textbf{343M}  \\ \midrule
Uniform & 68.30                                                                          & {1.28M}  & {95.00}                                                    & {204M}  \\
Budget from \alg & \textbf{69.72}                                                                          & {1.33M}  & {94.80}                                                    & {182M}  \\
Budget from \alg & 68.01                                                                          & \textbf{1.24M}  & \textbf{95.15}                                                    & \textbf{162M}  \\

\bottomrule
\end{tabular}
}
\end{table}

Table~\ref{tab:res50budgettrans} shows the effectiveness/transferability  of the learnt non-uniform budget through \alg for 90\% sparse ResNet50 on ImageNet-1K using DNW~\citep{wortsman2019discovering}. DNW typically takes in a uniform sparsity budget and has an accuracy of 74\% for a 90\% sparse ResNet50. Using ERK non-uniform budget for 90\% sparsity results in a 0.1\% increase in accuracy at the cost 2.35$\times$ inference FLOPs. Training DNW with the learnt budget from \alg  results in a reduction of FLOPs by 66M (16\%) while maintaining accuracy. In the 95\% sparsity regime, the learnt budget can improve the accuracy of DNW by up to 1.42\% over uniform along with a reduction in FLOPs by at least 22M (11\%).

\begin{table}[!ht]
\caption{Effect of various layer-wise sparsity budgets when used with GMP for ResNet50 on ImageNet-1K.}
\label{tab:gmpbt}
\resizebox{\columnwidth}{!}{
\begin{tabular}{@{}lcccc@{}}
\toprule
Method                          & \multicolumn{1}{c}{\begin{tabular}[c]{@{}c@{}}Top-1 Acc \\ (\%)\end{tabular}} & Params & \begin{tabular}[c]{@{}c@{}}Sparsity \\ (\%)\end{tabular} & FLOPs \\ \midrule
Uniform & 73.91                                                                          & {2.56M}  & {90.00}                                                    & {409M}  \\
Budget from \alg & \textbf{74.13}                                                                          & \textbf{2.49M}  & \textbf{90.23}                                                    & \textbf{343M}  \\\midrule
Uniform & 57.90                                                                          & {0.51M}  & {98.00}                                                    & {82M}  \\
Budget from \alg & \textbf{59.47}                                                                          & \textbf{0.50M}  & \textbf{98.05}                                                    & \textbf{73M}  \\ \bottomrule
\end{tabular}
}
\end{table}

Similarly, these budgets can also be used for other methods like GMP~\citep{zhu2017prune}. Table~\ref{tab:gmpbt} shows that the learnt sparsity budgets can lead to an increase in accuracy by 0.22\% and 1.57\% in 90\% and 98\% sparsity regimes respectively when used with GMP. Accuracy gains over uniform sparsity are also accompanied by a significant reduction in inference FLOPs. Note that the learnt non-uniform sparsity budgets can also be obtained using smaller representative datasets instead of expensive large-scale experiments.

The major drawback of \alg is the tuning of the weight-decay parameter, $\lambda$ and finer-tuning with $s_{\rm init}$ to obtain the targeted overall sparsity. One way to circumvent this issue is to freeze the non-uniform sparsity distribution in the middle of training when the overall sparsity constraints are met and train for the remaining epochs. This might not potentially give the best results but can give a similar budget which can be then transferred to methods like GMP or DNW. Another drawback of \alg is the function $g$ for the threshold. The stability, expressivity, and sparsification capability of \alg depends on $g$. However, it should be noted that sigmoid and exponential functions work just fine, as $g$, for \alg.  
\section{Conclusions}
\label{sec:conc}
This paper proposed Soft Threshold Reparameterization (\alg), a novel use of the soft-threshold operator, for the weights in DNN, to smoothly induce sparsity while learning layer-wise pruning thresholds thereby obtaining a non-uniform sparsity budget. Extensive experimentation showed that \alg is state-of-the-art for unstructured sparsity in CNNs for ImageNet-1K while also being effective for structured sparsity in RNNs. Our method results in sparse models that have significantly lesser inference costs than the baselines. In particular, \alg achieves the same accuracy as the baselines for 90\% sparse MobileNetV1 with 50\% lesser FLOPs. \alg has $\sim$10\% higher accuracy than the existing methods in ultra sparse (99\%) regime for ResNet50 showing the effectiveness of the learnt non-uniform sparsity budgets. \alg can also induce low-rank structure in RNNs while increasing the prediction accuracy showing the generalizability of the proposed reparameterization. Finally, \alg is easy to adapt and the learnt budgets are transferable.

\section*{Acknowledgments}
We are grateful to Keivan Alizadeh, Tapan Chugh, Tim Dettmers, Erich Elsen, Utku Evci, Daniel Gordon, Gabriel Ilharco, Sarah Pratt, James Park, Mohammad Rastegari and Matt Wallingford  for helpful discussions and feedback. Mitchell Wortsman is in part supported by AI2 Fellowship in AI. Sham Kakade acknowledges funding from the Washington Research Foundation for Innovation in Data-intensive Discovery, and the NSF Awards CCF-1637360, CCF-1703574, and CCF-1740551. Ali Farhadi acknowledges funding from the NSF Awards IIS 1652052, IIS 17303166, DARPA N66001-19-2-4031, 67102239 and gifts from Allen Institute for Artificial Intelligence.

\bibliography{local}
\clearpage
\newpage
\onecolumn

\appendix

\section{Appendix}

\subsection{Characterization of $g$}
\label{sec:g}
For the training dynamics of $s$, we propose some desired properties for choosing $g:\R \to\R_{++}$:
\begin{itemize}
    \item $0<g(s)$, $\lim\limits_{s\to -\infty} g(s) = 0$, and $\lim\limits_{s\to\infty}g(s) = \infty$.
    \item $\exists\ G\in\R_{++} \ni 0<g'(s)\leq G\ \forall\ s\in\R$.
    \item $g'(s_{\rm init}) < 1$ providing us a handle on the dynamics of $s$.
\end{itemize}
For simplicity, the choice of $g$ were the logistic sigmoid function, $g(s) = \frac{k}{1+e^{-s}}$, and the exponential function,  $g(s) = ke^{s}$, for $k\in\R_k$, since in most of the experimental scenarios, we almost always have $s<0$ throughout the training making it satisfy all the desired properties in $\R_{-}$. One can choose $k$ as an appropriate scaling factor based on the final weight distribution of a given DNN. All the CNN experiments in this paper we use the logistic sigmoid function with $k=1$, as the weights' final learnt values are typically $\ll 1$, and low-rank RNN use the exponential function with $k=1$. It should be noted that better functional choices might exist for $g$ and can affect the expressivity and dynamics of \alg parameterization for inducing sparsity.

\subsection{Gradient w.r.t. $\{s_l\}_{l\in\squarebrack{L}}$}
\label{sec:sl_update}
The gradient of $s_l\ \forall\ l\in\squarebrack{L}$ takes an even interesting form
\begin{align}
    \nabla_{s_l}\Lcal \big(\widetilde\W_l(s_l)\big)\nonumber &= \nabla_{s_l}\Lcal \round{\Scal_g(\W_l,s_l)}\nonumber\\
    &= -g^\prime(s_l)\Pcal\round{\W_l,g(s_l)}
\end{align}
Where $\Pcal\round{\W_l,g(s_l)} := \inner{\nabla_{\widetilde\W_l(s_l)}\Lcal \big(\widetilde\W(s_l)\big),\sign{\W_l}\odot\indicator{\widetilde\W_l(s_l)\neq 0}}$. Thus the final update equation for $s_l\ \forall\ l\in\squarebrack{L}$ becomes
\begin{eqnarray}
    s_l^{(t+1)} \gets s_l^{(t)} + \eta_t g^\prime(s_l^{(t)})\Pcal\round{\W_l^{(t)},g\round{s_l^{(t)}}} - \eta_t\lambda s_l^{(t)}
\end{eqnarray}
where $\lambda$ is its $\ell_2$ regularization hyperparameter.

\begin{algorithm}[!ht]
\caption{PyTorch code for \ensuremath{{\rm STRConv}} with per-layer threshold.}
\label{alg:code}

\definecolor{codeblue}{rgb}{0.25,0.5,0.5}
\definecolor{codeblue2}{rgb}{0,0,1}
\lstset{
  backgroundcolor=\color{white},
  basicstyle=\fontsize{7.2pt}{7.2pt}\ttfamily\selectfont,
  columns=fullflexible,
  breaklines=true,
  captionpos=b,
  commentstyle=\fontsize{7.2pt}{7.2pt}\color{codeblue},
  keywordstyle=\fontsize{7.2pt}{7.2pt}\color{codeblue2},
}
\begin{lstlisting}[language=python]
import torch
import torch.nn as nn
import torch.nn.functional as F

from args import args as parser_args

def softThreshold(x, s, g=torch.sigmoid):
    # STR on a weight x (can be a tensor) with "s" (typically a scalar, but can be a tensor) with function "g".
    return torch.sign(x)*torch.relu(torch.abs(x)-g(s))

class STRConv(nn.Conv2d): # Overloaded Conv2d which can replace nn.Conv2d
    def __init__(self, *args, **kwargs):
        super().__init__(*args, **kwargs)
        # "g" can be chosen appropriately, but torch.sigmoid works fine. 
        self.g = torch.sigmoid
        # parser_args gets arguments from command line. sInitValue is the initialization of "s" for all layers. It can take in different values per-layer as well.
        self.s = nn.Parameter(parser_args.sInitValue*torch.ones([1, 1]))
        # "s" can be per-layer (a scalar), global (a shared scalar across layers), per-channel/filter (a vector) or per individual weight (a tensor of the size self.weight). All the experiments use per-layer "s" (a scalar) in the paper.
    
    def forward(self, x):
        
        sparseWeight = softThreshold(self.weight, self.s, self.g)
        # Parameters except "x" and "sparseWeight" can be chosen appropriately. All the experiments use default PyTorch arguments. 
        x = F.conv2d(x, sparseWeight, self.bias, self.stride, self.padding, self.dilation, self.groups)
        
        return x

# FC layer is implemented as a 1x1 Conv2d and STRConv is used for FC layer as well.
\end{lstlisting}
\end{algorithm}

\subsection{ResNet50 Learnt Budgets and Backbone Sparsities}
\label{sec:res50budget}
Table~\ref{tab:res50big} lists the non-uniform sparsity budgets learnt through \alg across the sparsity regimes of 80\%, 90\%, 95\%, 96.5\%, 98\% and 99\% for ResNet50 on ImageNet-1K. The table also lists the backbone sparsities of every budget. It is clear that \alg results in a higher than expected sparsity in the backbones of CNNs resulting in efficient backbones for transfer learning.

Table~\ref{tab:res50nub} summarizes all the sparsity budgets for 90\% sparse ResNet50 on ImageNet-1K obtained using various methods. This table also shows that the backbone sparsities learnt through \alg are considerably higher than that of the baselines.

One can use these budgets directly for techniques like GMP and DNW for a variety of datasets and have significant accuracy gains as shown in the Table~\ref{tab:res50budgettrans}.

\begin{table}[!ht]
\centering
\caption{The non-uniform sparsity budgets for various sparsity ranges learnt through \alg for ResNet50 on ImageNet-1K. FLOPs distribution per layer can be computed as $\frac{100-s_i}{100}*\text{FLOPs}_i$, where $s_i$ and $\text{FLOPs}_i$ are the sparsity and FLOPs of the layer $i$.}
\label{tab:res50big}
\resizebox{\columnwidth}{!}{
\begin{tabular}{@{}l|rr|cccccccccccccccc@{}}
\toprule
Metric                           & \multicolumn{1}{c}{\begin{tabular}[c]{@{}c@{}}Fully Dense \\ Params\end{tabular}} & \multicolumn{1}{c|}{\begin{tabular}[c]{@{}c@{}}Fully Dense\\ FLOPs\end{tabular}} & \multicolumn{16}{c}{Sparsity (\%)}                                                                                           \\ \midrule
Overall                          & 25502912                                                                          & 4089284608                                                                       & 79.55 & 81.27 & 87.70 & 90.23 & 90.55 & 94.80 & 95.03 & 95.15 & 96.11 & 96.53 & 97.78 & 98.05 & 98.22 & 98.79 & 98.98 & 99.10 \\
Backbone                         & 23454912                                                                          & 4087136256                                                                       & 82.07 & 83.79 & 90.08 & 92.47 & 92.77 & 96.51 & 96.71 & 96.84 & 97.64 & 97.92 & 98.82 & 98.99 & 99.11 & 99.46 & 99.58 & 99.64 \\ \midrule
Layer 1 - conv1                  & 9408                                                                              & 118013952                                                                        & 51.46 & 51.40 & 63.02 & 59.80 & 59.83 & 64.87 & 67.36 & 66.96 & 72.11 & 69.46 & 73.29 & 73.47 & 72.05 & 75.12 & 76.12 & 77.75 \\
Layer 2 - layer1.0.conv1         & 4096                                                                              & 12845056                                                                         & 69.36 & 73.24 & 87.57 & 83.28 & 85.18 & 89.60 & 91.41 & 91.11 & 92.38 & 91.75 & 94.46 & 94.51 & 94.60 & 95.95 & 96.53 & 96.51 \\
Layer 3 - layer1.0.conv2         & 36864                                                                             & 115605504                                                                        & 77.85 & 76.26 & 90.87 & 89.48 & 87.31 & 94.79 & 94.27 & 95.04 & 95.69 & 96.07 & 97.36 & 97.77 & 98.35 & 98.51 & 98.59 & 98.84 \\
Layer 4 - layer1.0.conv3         & 16384                                                                             & 51380224                                                                         & 74.81 & 74.65 & 86.52 & 85.80 & 85.25 & 91.85 & 92.78 & 93.67 & 94.13 & 94.69 & 96.61 & 97.03 & 97.37 & 98.04 & 98.21 & 98.47 \\
Layer 5 - layer1.0.downsample.0  & 16384                                                                             & 51380224                                                                         & 70.95 & 72.96 & 83.53 & 83.34 & 82.56 & 89.13 & 90.62 & 90.17 & 91.83 & 92.69 & 95.48 & 94.89 & 95.68 & 96.98 & 97.56 & 97.72 \\
Layer 6 - layer1.1.conv1         & 16384                                                                             & 51380224                                                                         & 80.27 & 79.58 & 89.82 & 89.89 & 88.51 & 94.56 & 96.64 & 95.78 & 95.81 & 96.81 & 98.79 & 98.90 & 98.98 & 99.13 & 99.62 & 99.47 \\
Layer 7 - layer1.1.conv2         & 36864                                                                             & 115605504                                                                        & 81.36 & 80.95 & 91.75 & 90.60 & 89.61 & 94.70 & 95.78 & 96.18 & 96.42 & 97.26 & 98.65 & 99.07 & 99.40 & 99.11 & 99.31 & 99.56 \\
Layer 8 - layer1.1.conv3         & 16384                                                                             & 51380224                                                                         & 84.45 & 80.11 & 91.22 & 91.70 & 90.21 & 95.17 & 97.05 & 95.81 & 96.34 & 97.23 & 98.68 & 98.76 & 98.90 & 99.16 & 99.57 & 99.46 \\
Layer 9 - layer1.2.conv1         & 16384                                                                             & 51380224                                                                         & 78.23 & 79.79 & 90.12 & 88.07 & 89.36 & 94.62 & 95.94 & 94.74 & 96.23 & 96.75 & 97.96 & 98.41 & 98.72 & 99.38 & 99.35 & 99.46 \\
Layer 10 - layer1.2.conv2        & 36864                                                                             & 115605504                                                                        & 76.01 & 81.53 & 91.06 & 87.03 & 88.27 & 93.90 & 95.63 & 94.26 & 96.24 & 96.11 & 97.54 & 98.27 & 98.44 & 99.32 & 99.19 & 99.39 \\
Layer 11 - layer1.2.conv3        & 16384                                                                             & 51380224                                                                         & 84.47 & 83.28 & 94.95 & 90.99 & 92.64 & 95.76 & 96.95 & 96.01 & 96.87 & 97.31 & 98.38 & 98.60 & 98.72 & 99.38 & 99.27 & 99.51 \\
Layer 12 - layer2.0.conv1        & 32768                                                                             & 102760448                                                                        & 73.74 & 73.96 & 86.78 & 85.95 & 85.90 & 92.32 & 94.79 & 93.86 & 94.62 & 95.64 & 97.19 & 98.22 & 98.52 & 98.48 & 98.84 & 98.92 \\
Layer 13 - layer2.0.conv2        & 147456                                                                            & 115605504                                                                        & 82.56 & 85.70 & 91.31 & 93.91 & 94.03 & 97.54 & 97.43 & 97.65 & 98.38 & 98.62 & 99.24 & 99.23 & 99.40 & 99.61 & 99.67 & 99.63 \\
Layer 14 - layer2.0.conv3        & 65536                                                                             & 51380224                                                                         & 84.70 & 83.55 & 93.04 & 93.13 & 92.13 & 96.61 & 97.37 & 97.21 & 97.59 & 98.14 & 98.80 & 98.95 & 99.18 & 99.29 & 99.47 & 99.43 \\
Layer 15 - layer2.0.downsample.0 & 131072                                                                            & 102760448                                                                        & 85.10 & 87.66 & 92.78 & 94.96 & 95.13 & 98.07 & 97.97 & 98.15 & 98.70 & 98.88 & 99.37 & 99.35 & 99.40 & 99.69 & 99.68 & 99.71 \\
Layer 16 - layer2.1.conv1        & 65536                                                                             & 51380224                                                                         & 85.42 & 85.79 & 94.04 & 95.31 & 94.94 & 97.92 & 98.53 & 98.21 & 98.84 & 99.06 & 99.46 & 99.53 & 99.72 & 99.78 & 99.81 & 99.80 \\
Layer 17 - layer2.1.conv2        & 147456                                                                            & 115605504                                                                        & 76.95 & 82.75 & 87.63 & 91.50 & 91.76 & 95.59 & 97.22 & 96.07 & 97.32 & 97.80 & 98.24 & 98.24 & 98.60 & 99.24 & 99.66 & 99.33 \\
Layer 18 - layer2.1.conv3        & 65536                                                                             & 51380224                                                                         & 84.76 & 84.71 & 93.10 & 93.66 & 93.23 & 97.00 & 98.18 & 97.35 & 98.06 & 98.41 & 98.96 & 99.21 & 99.32 & 99.55 & 99.58 & 99.59 \\
Layer 19 - layer2.2.conv1        & 65536                                                                             & 51380224                                                                         & 84.30 & 85.34 & 92.70 & 94.61 & 94.76 & 97.72 & 97.91 & 98.21 & 98.54 & 98.98 & 99.24 & 99.35 & 99.50 & 99.62 & 99.63 & 99.77 \\
Layer 20 - layer2.2.conv2        & 147456                                                                            & 115605504                                                                        & 84.28 & 85.43 & 92.99 & 94.86 & 94.90 & 97.52 & 97.21 & 98.11 & 98.19 & 99.04 & 99.28 & 99.37 & 99.46 & 99.63 & 99.59 & 99.72 \\
Layer 21 - layer2.2.conv3        & 65536                                                                             & 51380224                                                                         & 82.19 & 84.21 & 91.12 & 93.38 & 93.53 & 96.89 & 97.14 & 97.59 & 97.77 & 98.66 & 98.96 & 99.15 & 99.25 & 99.49 & 99.51 & 99.57 \\
Layer 22 - layer2.3.conv1        & 65536                                                                             & 51380224                                                                         & 83.37 & 84.41 & 90.46 & 93.26 & 93.50 & 96.71 & 97.89 & 96.99 & 98.14 & 98.36 & 99.10 & 99.23 & 99.33 & 99.53 & 99.75 & 99.60 \\
Layer 23 - layer2.3.conv2        & 147456                                                                            & 115605504                                                                        & 82.83 & 84.03 & 91.44 & 93.21 & 93.25 & 96.83 & 98.02 & 96.96 & 98.45 & 98.30 & 98.97 & 99.06 & 99.26 & 99.31 & 99.81 & 99.68 \\
Layer 24 - layer2.3.conv3        & 65536                                                                             & 51380224                                                                         & 82.93 & 85.65 & 91.02 & 94.14 & 93.56 & 97.20 & 97.97 & 97.04 & 98.16 & 98.36 & 98.88 & 98.97 & 99.20 & 99.32 & 99.67 & 99.62 \\
Layer 25 - layer3.0.conv1        & 131072                                                                            & 102760448                                                                        & 76.63 & 77.98 & 85.99 & 88.85 & 88.60 & 94.26 & 95.07 & 94.97 & 96.21 & 96.59 & 97.75 & 98.04 & 98.30 & 98.72 & 99.11 & 99.06 \\
Layer 26 - layer3.0.conv2        & 589824                                                                            & 115605504                                                                        & 87.35 & 88.68 & 94.39 & 96.14 & 96.19 & 98.51 & 98.77 & 98.72 & 99.11 & 99.23 & 99.53 & 99.59 & 99.64 & 99.73 & 99.80 & 99.81 \\
Layer 27 - layer3.0.conv3        & 262144                                                                            & 51380224                                                                         & 81.22 & 83.22 & 90.58 & 93.19 & 93.05 & 96.82 & 97.38 & 97.32 & 97.98 & 98.28 & 98.88 & 99.03 & 99.16 & 99.39 & 99.55 & 99.53 \\
Layer 28 - layer3.0.downsample.0 & 524288                                                                            & 102760448                                                                        & 89.75 & 90.99 & 96.05 & 97.20 & 97.16 & 98.96 & 99.21 & 99.20 & 99.50 & 99.58 & 99.78 & 99.82 & 99.86 & 99.91 & 99.94 & 99.93 \\
Layer 29 - layer3.1.conv1        & 262144                                                                            & 51380224                                                                         & 85.88 & 87.35 & 93.43 & 95.36 & 96.12 & 98.64 & 98.77 & 98.87 & 99.22 & 99.33 & 99.64 & 99.67 & 99.72 & 99.82 & 99.88 & 99.84 \\
Layer 30 - layer3.1.conv2        & 589824                                                                            & 115605504                                                                        & 85.06 & 86.24 & 92.74 & 95.06 & 95.30 & 98.09 & 98.28 & 98.36 & 98.75 & 99.08 & 99.46 & 99.48 & 99.54 & 99.69 & 99.76 & 99.76 \\
Layer 31 - layer3.1.conv3        & 262144                                                                            & 51380224                                                                         & 84.34 & 86.79 & 92.15 & 94.84 & 94.90 & 97.75 & 98.15 & 98.11 & 98.56 & 98.94 & 99.30 & 99.36 & 99.45 & 99.65 & 99.79 & 99.70 \\
Layer 32 - layer3.2.conv1        & 262144                                                                            & 51380224                                                                         & 87.51 & 89.15 & 94.15 & 96.77 & 96.46 & 98.81 & 98.83 & 98.96 & 99.19 & 99.44 & 99.67 & 99.71 & 99.74 & 99.82 & 99.85 & 99.89 \\
Layer 33 - layer3.2.conv2        & 589824                                                                            & 115605504                                                                        & 87.15 & 88.67 & 94.09 & 95.59 & 96.14 & 98.86 & 98.69 & 98.91 & 99.21 & 99.20 & 99.64 & 99.72 & 99.76 & 99.85 & 99.84 & 99.90 \\
Layer 34 - layer3.2.conv3        & 262144                                                                            & 51380224                                                                         & 84.86 & 86.90 & 92.40 & 94.99 & 94.99 & 98.19 & 98.19 & 98.42 & 98.76 & 98.97 & 99.42 & 99.56 & 99.62 & 99.76 & 99.75 & 99.88 \\
Layer 35 - layer3.3.conv1        & 262144                                                                            & 51380224                                                                         & 86.62 & 89.46 & 94.06 & 96.08 & 95.88 & 98.70 & 98.71 & 98.77 & 99.01 & 99.27 & 99.58 & 99.66 & 99.69 & 99.83 & 99.87 & 99.87 \\
Layer 36 - layer3.3.conv2        & 589824                                                                            & 115605504                                                                        & 86.52 & 87.97 & 93.56 & 96.10 & 96.11 & 98.70 & 98.82 & 98.89 & 99.19 & 99.31 & 99.68 & 99.73 & 99.77 & 99.88 & 99.87 & 99.93 \\
Layer 37 - layer3.3.conv3        & 262144                                                                            & 51380224                                                                         & 84.19 & 86.81 & 92.32 & 94.94 & 94.91 & 98.20 & 98.37 & 98.43 & 98.82 & 99.00 & 99.51 & 99.57 & 99.64 & 99.81 & 99.81 & 99.87 \\
Layer 38 - layer3.4.conv1        & 262144                                                                            & 51380224                                                                         & 85.85 & 88.40 & 93.55 & 95.49 & 95.86 & 98.35 & 98.44 & 98.55 & 98.79 & 98.96 & 99.54 & 99.59 & 99.60 & 99.82 & 99.86 & 99.87 \\
Layer 39 - layer3.4.conv2        & 589824                                                                            & 115605504                                                                        & 85.96 & 87.38 & 93.27 & 95.66 & 95.63 & 98.41 & 98.58 & 98.56 & 99.19 & 99.26 & 99.64 & 99.69 & 99.67 & 99.87 & 99.90 & 99.92 \\
Layer 40 - layer3.4.conv3        & 262144                                                                            & 51380224                                                                         & 83.45 & 85.76 & 91.75 & 94.49 & 94.35 & 97.67 & 98.09 & 97.99 & 98.65 & 98.94 & 99.49 & 99.52 & 99.48 & 99.77 & 99.86 & 99.85 \\
Layer 41 - layer3.5.conv1        & 262144                                                                            & 51380224                                                                         & 83.33 & 85.77 & 91.79 & 95.09 & 94.24 & 97.46 & 97.89 & 97.92 & 98.71 & 98.90 & 99.35 & 99.52 & 99.58 & 99.76 & 99.79 & 99.83 \\
Layer 42 - layer3.5.conv2        & 589824                                                                            & 115605504                                                                        & 84.98 & 86.67 & 92.48 & 94.92 & 95.13 & 97.88 & 98.14 & 98.32 & 98.91 & 99.00 & 99.44 & 99.58 & 99.69 & 99.80 & 99.83 & 99.87 \\
Layer 43 - layer3.5.conv3        & 262144                                                                            & 51380224                                                                         & 79.78 & 82.23 & 89.39 & 93.14 & 92.76 & 96.59 & 97.04 & 97.30 & 98.10 & 98.41 & 99.03 & 99.25 & 99.44 & 99.61 & 99.71 & 99.75 \\
Layer 44 - layer4.0.conv1        & 524288                                                                            & 102760448                                                                        & 77.83 & 79.61 & 87.11 & 90.32 & 90.64 & 95.39 & 95.84 & 95.92 & 97.17 & 97.35 & 98.36 & 98.60 & 98.83 & 99.20 & 99.37 & 99.42 \\
Layer 45 - layer4.0.conv2        & 2359296                                                                           & 115605504                                                                        & 86.18 & 88.00 & 93.53 & 95.66 & 95.78 & 98.31 & 98.47 & 98.55 & 99.08 & 99.16 & 99.54 & 99.63 & 99.69 & 99.81 & 99.85 & 99.86 \\
Layer 46 - layer4.0.conv3        & 1048576                                                                           & 51380224                                                                         & 78.43 & 80.48 & 87.85 & 91.14 & 91.27 & 96.00 & 96.40 & 96.47 & 97.53 & 97.92 & 98.81 & 99.00 & 99.15 & 99.45 & 99.57 & 99.61 \\
Layer 47 - layer4.0.downsample.0 & 2097152                                                                           & 102760448                                                                        & 88.49 & 89.98 & 95.03 & 96.79 & 96.90 & 98.91 & 99.06 & 99.11 & 99.45 & 99.51 & 99.77 & 99.82 & 99.85 & 99.92 & 99.94 & 99.94 \\
Layer 48 - layer4.1.conv1        & 1048576                                                                           & 51380224                                                                         & 82.07 & 84.02 & 90.34 & 93.69 & 93.72 & 97.15 & 97.56 & 97.76 & 98.45 & 98.75 & 99.27 & 99.36 & 99.54 & 99.67 & 99.76 & 99.80 \\
Layer 49 - layer4.1.conv2        & 2359296                                                                           & 115605504                                                                        & 83.42 & 85.23 & 91.16 & 93.98 & 93.93 & 97.26 & 97.58 & 97.71 & 98.36 & 98.67 & 99.25 & 99.34 & 99.50 & 99.68 & 99.76 & 99.80 \\
Layer 50 - layer4.1.conv3        & 1048576                                                                           & 51380224                                                                         & 78.08 & 79.96 & 86.66 & 90.48 & 90.22 & 95.22 & 95.76 & 95.89 & 96.88 & 97.65 & 98.70 & 98.85 & 99.13 & 99.45 & 99.58 & 99.66 \\
Layer 51 - layer4.2.conv1        & 1048576                                                                           & 51380224                                                                         & 76.34 & 77.93 & 84.98 & 87.57 & 88.47 & 93.90 & 93.87 & 94.16 & 95.55 & 95.91 & 97.66 & 97.97 & 98.15 & 98.88 & 99.08 & 99.22 \\
Layer 52 - layer4.2.conv2        & 2359296                                                                           & 115605504                                                                        & 73.57 & 74.97 & 82.32 & 84.37 & 86.01 & 91.92 & 91.66 & 92.22 & 94.02 & 94.16 & 96.65 & 97.13 & 97.29 & 98.44 & 98.74 & 99.00 \\
Layer 53 - layer4.2.conv3        & 1048576                                                                           & 51380224                                                                         & 68.78 & 70.38 & 78.11 & 80.29 & 81.73 & 89.64 & 89.43 & 89.65 & 91.40 & 92.65 & 96.02 & 96.72 & 96.93 & 98.47 & 98.83 & 99.15 \\
Layer 54 - fc                    & 2048000                                                                           & 2048000                                                                          & 50.65 & 52.46 & 60.48 & 64.50 & 65.12 & 75.20 & 75.73 & 75.80 & 78.57 & 80.69 & 85.96 & 87.26 & 88.03 & 91.11 & 92.15 & 92.87 \\ 
AP - adaptive average pool before fc &0&100352&~0.00&~0.00&~0.00&~0.00&~0.00&~0.00&~0.00&~0.00&~0.00&~0.00&~0.00&~0.00&~0.00&~0.00&~0.00&~0.00\\\bottomrule
\end{tabular}}
\end{table}

\begin{table}[!ht]
\centering
\caption{The non-uniform sparsity budgets learnt multiple methods for 90\% sparse ResNet50 on ImageNet-1K. FLOPs distribution per layer can be computed as $\frac{100-s_i}{100}*\text{FLOPs}_i$, where $s_i$ and $\text{FLOPs}_i$ are the sparsity and FLOPs of the layer $i$.}
\label{tab:res50nub}
\resizebox{0.6\columnwidth}{0.45\columnwidth}{
\begin{tabular}{@{}l|rr|cccccc@{}}
\toprule
\multirow{2}{*}{Metric}          & \multicolumn{1}{c}{\multirow{2}{*}{\begin{tabular}[c]{@{}c@{}}Fully Dense \\ Params\end{tabular}}} & \multicolumn{1}{c|}{\multirow{2}{*}{\begin{tabular}[c]{@{}c@{}}Fully Dense \\ FLOPs\end{tabular}}} & \multicolumn{5}{c}{Sparsity (\%)}                     \\ \cmidrule(l){4-9} 
                                 & \multicolumn{1}{c}{}                                                                               & \multicolumn{1}{c|}{}                                                                              & \alg & Uniform & ERK   & SNFS  & VD  &  GS  \\ \midrule
Overall                          & 25502912                                                                                           & 4089284608                                                                                        & 90.23               & 90.00   & 90.07 & 90.06 & 90.27 & 89.54 \\
Backbone                         & 23454912                                                                                           & 4087136256                                                                                        & 92.47               & 90.00   & 89.82 & 89.44 & 91.41 & 90.95 \\ \midrule
Layer 1 - conv1                  & 9408                                                                                               & 118013952                                                                                         & 59.80               & 90.00   & 58.00 & 2.50  & 31.39 & 35.11 \\
Layer 2 - layer1.0.conv1         & 4096                                                                                               & 12845056                                                                                          & 83.28               & 90.00   & 0.00  & 2.50  & 39.50 & 56.05 \\
Layer 3 - layer1.0.conv2         & 36864                                                                                              & 115605504                                                                                         & 89.48               & 90.00   & 82.00 & 2.50  & 67.87 & 75.04 \\
Layer 4 - layer1.0.conv3         & 16384                                                                                              & 51380224                                                                                          & 85.80               & 90.00   & 4.00  & 2.50  & 64.87 & 70.31\\
Layer 5 - layer1.0.downsample.0  & 16384                                                                                              & 51380224                                                                                          & 83.34               & 90.00   & 4.00  & 2.50  & 60.38 & 66.88\\
Layer 6 - layer1.1.conv1         & 16384                                                                                              & 51380224                                                                                          & 89.89               & 90.00   & 4.00  & 2.50  & 61.35 & 75.09\\
Layer 7 - layer1.1.conv2         & 36864                                                                                              & 115605504                                                                                         & 90.60               & 90.00   & 82.00 & 2.50  & 64.38 & 80.42\\
Layer 8 - layer1.1.conv3         & 16384                                                                                              & 51380224                                                                                          & 91.70               & 90.00   & 4.00  & 2.50  & 65.83 & 80.00\\
Layer 9 - layer1.2.conv1         & 16384                                                                                              & 51380224                                                                                          & 88.07               & 90.00   & 4.00  & 2.50  & 68.75 & 75.21\\
Layer 10 - layer1.2.conv2        & 36864                                                                                              & 115605504                                                                                         & 87.03               & 90.00   & 82.00 & 2.50  & 70.86 & 74.95\\
Layer 11 - layer1.2.conv3        & 16384                                                                                              & 51380224                                                                                          & 90.99               & 90.00   & 4.00  & 2.50  & 54.05 & 79.28\\
Layer 12 - layer2.0.conv1        & 32768                                                                                              & 102760448                                                                                         & 85.95               & 90.00   & 43.00 & 2.50  & 57.10 & 70.89\\
Layer 13 - layer2.0.conv2        & 147456                                                                                             & 115605504                                                                                         & 93.91               & 90.00   & 91.00 & 62.90 & 78.65 & 85.39\\
Layer 14 - layer2.0.conv3        & 65536                                                                                              & 51380224                                                                                          & 93.13               & 90.00   & 52.00 & 11.00 & 85.49 & 83.54\\
Layer 15 - layer2.0.downsample.0 & 131072                                                                                             & 102760448                                                                                         & 94.96               & 90.00   & 71.00 & 66.10 & 79.96 & 88.36\\
Layer 16 - layer2.1.conv1        & 65536                                                                                              & 51380224                                                                                          & 95.31               & 90.00   & 52.00 & 32.60 & 72.07 & 88.25\\
Layer 17 - layer2.1.conv2        & 147456                                                                                             & 115605504                                                                                         & 91.50               & 90.00   & 91.00 & 61.60 & 84.41 & 85.37\\
Layer 18 - layer2.1.conv3        & 65536                                                                                              & 51380224                                                                                          & 93.66               & 90.00   & 52.00 & 20.80 & 79.19 & 86.53\\
Layer 19 - layer2.2.conv1        & 65536                                                                                              & 51380224                                                                                          & 94.61               & 90.00   & 52.00 & 29.10 & 73.94 & 86.40\\
Layer 20 - layer2.2.conv2        & 147456                                                                                             & 115605504                                                                                         & 94.86               & 90.00   & 91.00 & 63.90 & 78.48 & 88.29\\
Layer 21 - layer2.2.conv3        & 65536                                                                                              & 51380224                                                                                          & 93.38               & 90.00   & 52.00 & 22.90 & 78.09 & 85.87\\
Layer 22 - layer2.3.conv1        & 65536                                                                                              & 51380224                                                                                          & 93.26               & 90.00   & 52.00 & 27.60 & 78.66 & 84.87\\
Layer 23 - layer2.3.conv2        & 147456                                                                                             & 115605504                                                                                         & 93.21               & 90.00   & 91.00 & 65.30 & 84.38 & 87.14\\
Layer 24 - layer2.3.conv3        & 65536                                                                                              & 51380224                                                                                          & 94.14               & 90.00   & 52.00 & 25.70 & 82.07 & 86.84\\
Layer 25 - layer3.0.conv1        & 131072                                                                                             & 102760448                                                                                         & 88.85               & 90.00   & 71.00 & 48.70 & 66.56 & 78.40\\
Layer 26 - layer3.0.conv2        & 589824                                                                                             & 115605504                                                                                         & 96.14               & 90.00   & 96.00 & 90.20 & 87.92 & 92.93\\
Layer 27 - layer3.0.conv3        & 262144                                                                                             & 51380224                                                                                          & 93.19               & 90.00   & 76.00 & 73.30 & 92.19 & 86.19\\
Layer 28 - layer3.0.downsample.0 & 524288                                                                                             & 102760448                                                                                         & 97.20               & 90.00   & 86.00 & 93.70 & 88.76 & 94.66\\
Layer 29 - layer3.1.conv1        & 262144                                                                                             & 51380224                                                                                          & 95.36               & 90.00   & 76.00 & 81.10 & 91.79 & 93.60\\
Layer 30 - layer3.1.conv2        & 589824                                                                                             & 115605504                                                                                         & 95.06               & 90.00   & 96.00 & 90.40 & 92.47 & 93.07\\
Layer 31 - layer3.1.conv3        & 262144                                                                                             & 51380224                                                                                          & 94.84               & 90.00   & 76.00 & 78.10 & 88.88 & 90.54\\
Layer 32 - layer3.2.conv1        & 262144                                                                                             & 51380224                                                                                          & 96.77               & 90.00   & 76.00 & 80.40 & 84.86 & 93.44\\
Layer 33 - layer3.2.conv2        & 589824                                                                                             & 115605504                                                                                         & 95.59               & 90.00   & 96.00 & 90.80 & 91.50 & 93.73\\
Layer 34 - layer3.2.conv3        & 262144                                                                                             & 51380224                                                                                          & 94.99               & 90.00   & 76.00 & 79.30 & 81.59 & 91.13\\
Layer 35 - layer3.3.conv1        & 262144                                                                                             & 51380224                                                                                          & 96.08               & 90.00   & 76.00 & 80.70 & 76.64 & 93.18\\
Layer 36 - layer3.3.conv2        & 589824                                                                                             & 115605504                                                                                         & 96.10               & 90.00   & 96.00 & 90.70 & 91.26 & 93.63\\
Layer 37 - layer3.3.conv3        & 262144                                                                                             & 51380224                                                                                          & 94.94               & 90.00   & 76.00 & 79.00 & 85.46 & 91.63\\
Layer 38 - layer3.4.conv1        & 262144                                                                                             & 51380224                                                                                          & 95.49               & 90.00   & 76.00 & 79.40 & 85.33 & 91.98\\
Layer 39 - layer3.4.conv2        & 589824                                                                                             & 115605504                                                                                         & 95.66               & 90.00   & 96.00 & 91.00 & 91.57 & 94.21\\
Layer 40 - layer3.4.conv3        & 262144                                                                                             & 51380224                                                                                          & 94.49               & 90.00   & 76.00 & 79.00 & 86.19 & 91.63\\
Layer 41 - layer3.5.conv1        & 262144                                                                                             & 51380224                                                                                          & 95.09               & 90.00   & 76.00 & 78.30 & 84.64 & 90.72\\
Layer 42 - layer3.5.conv2        & 589824                                                                                             & 115605504                                                                                         & 94.92               & 90.00   & 96.00 & 91.00 & 91.14 & 93.43\\
Layer 43 - layer3.5.conv3        & 262144                                                                                             & 51380224                                                                                          & 93.14               & 90.00   & 76.00 & 78.20 & 84.09 & 89.56\\
Layer 44 - layer4.0.conv1        & 524288                                                                                             & 102760448                                                                                         & 90.32               & 90.00   & 86.00 & 85.80 & 77.90 & 85.35\\
Layer 45 - layer4.0.conv2        & 2359296                                                                                            & 115605504                                                                                         & 95.66               & 90.00   & 98.00 & 97.60 & 96.53 & 95.07\\
Layer 46 - layer4.0.conv3        & 1048576                                                                                            & 51380224                                                                                          & 91.14               & 90.00   & 88.00 & 93.20 & 93.52 & 89.21\\
Layer 47 - layer4.0.downsample.0 & 2097152                                                                                            & 102760448                                                                                         & 96.79               & 90.00   & 93.00 & 98.80 & 93.80 & 96.72\\
Layer 48 - layer4.1.conv1        & 1048576                                                                                            & 51380224                                                                                          & 93.69               & 90.00   & 88.00 & 94.10 & 94.96 & 92.69\\
Layer 49 - layer4.1.conv2        & 2359296                                                                                            & 115605504                                                                                         & 93.98               & 90.00   & 98.00 & 97.70 & 97.76 & 93.85\\
Layer 50 - layer4.1.conv3        & 1048576                                                                                            & 51380224                                                                                          & 90.48               & 90.00   & 88.00 & 94.20 & 94.53 & 89.84\\
Layer 51 - layer4.2.conv1        & 1048576                                                                                            & 51380224                                                                                          & 87.57               & 90.00   & 88.00 & 93.60 & 94.19 & 85.91\\
Layer 52 - layer4.2.conv2        & 2359296                                                                                            & 115605504                                                                                         & 84.37               & 90.00   & 98.00 & 97.90 & 94.92 & 87.14\\
Layer 53 - layer4.2.conv3        & 1048576                                                                                            & 51380224                                                                                          & 80.29               & 90.00   & 88.00 & 94.50 & 89.64 & 80.65\\
Layer 54 - fc                    & 2048000                                                                                            & 2048000                                                                                           & 64.50               & 90.00   & 93.00 & 97.10 & 77.17 & 73.43 \\ 
AP - adaptive average pool before fc &0&100352&~0.00&~0.00&~0.00&~0.00&~0.00&~0.00\\\bottomrule
\end{tabular}
}
\end{table}
\subsection{MobileNetV1 Sparsity and FLOPs Budget Distributions}
\label{sec:mbv1dist}
Table~\ref{tab:mbv1nub} summarizes all the sparsity budgets for 90\% sparse MobileNetV1 on ImageNet-1K obtained using various methods. Note that GMP here makes the first and depthwise (dw) convolution layers dense, hence it is not the standard uniform sparsity. This table also shows that the backbone sparsities learnt through \alg are considerably higher than that of GMP.

\begin{figure}[!ht]
	\includegraphics[width=\columnwidth]{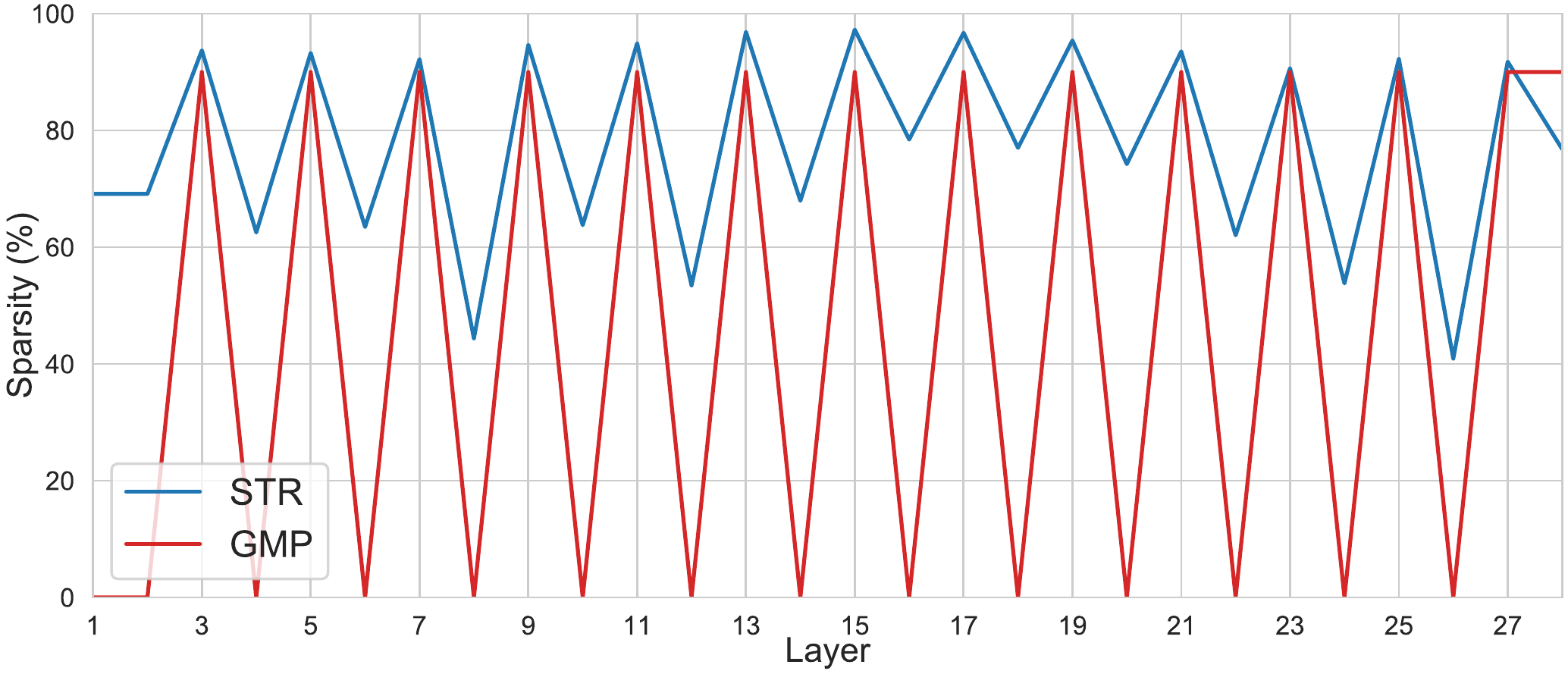}
	\caption{Layer-wise sparsity budget for the 90\% sparse MobileNetV1 models on ImageNet-1K using various sparsification techniques.}
 	\label{fig:sdmbv1}
\end{figure}

\begin{figure}[!ht]
	\includegraphics[width=\columnwidth]{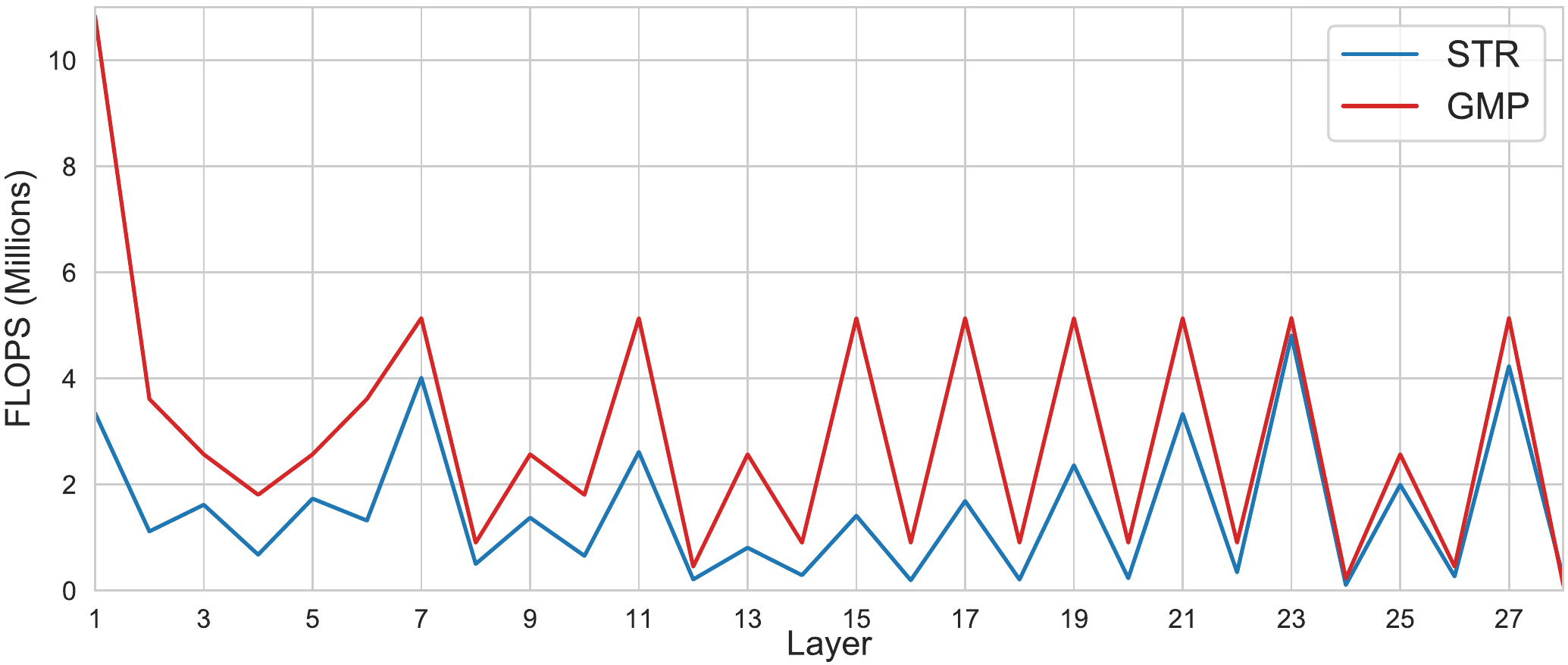}
	\caption{Layer-wise FLOPs distribution for the 90\% sparse MobileNetV1 models on ImageNet-1K using various sparsification techniques.}
 	\label{fig:fdmbv1}
\end{figure}

Figure~\ref{fig:sdmbv1} shows the sparsity distribution across layers when compared to GMP and Figure~\ref{fig:fdmbv1} shows the FLOPs distribution across layers when compared to GMP for 90\% sparse MobileNetV1 models on ImageNet-1K.

It is interesting to notice that \alg automatically keeps depthwise separable (the valleys in Figure~\ref{fig:sdmbv1}) convolution layers less sparse than the rest to maximize accuracy which is the reason GMP keeps them fully dense. 

\twocolumn

\begin{table}[!ht]
\centering
\caption{The non-uniform sparsity budgets learnt multiple methods for 90\% sparse MobileNetV1 on ImageNet-1K. FLOPs distribution per layer can be computed as $\frac{100-s_i}{100}*\text{FLOPs}_i$, where $s_i$ and $\text{FLOPs}_i$ are the sparsity and FLOPs of the layer $i$.}
\label{tab:mbv1nub}
\resizebox{\columnwidth}{!}{
\begin{tabular}{@{}l|rr|cc@{}}
\toprule
\multicolumn{1}{l|}{\multirow{2}{*}{Metric}} & \multicolumn{1}{c}{\multirow{2}{*}{\begin{tabular}[c]{@{}c@{}}Fully Dense \\ Params\end{tabular}}} & \multicolumn{1}{c|}{\multirow{2}{*}{\begin{tabular}[c]{@{}c@{}}Fully Dense \\ FLOPs\end{tabular}}} & \multicolumn{2}{l}{Sparsity (\%)} \\ \cmidrule(l){4-5} 
\multicolumn{1}{l|}{}                        & \multicolumn{1}{c}{}                                                                               & \multicolumn{1}{c|}{}                                                                              & \alg    & GMP      \\ \midrule
Overall                                      & 4209088                                                                                            & 568740352                                                                                          & 89.01                  & 89.03    \\
Backbone                                     & 3185088                                                                                            & 567716352                                                                                          & 92.93                  & 88.71    \\\midrule
Layer 1                                      & 864                                                                                                & 10838016                                                                                           & 69.10                  & 0.00     \\
Layer 2 (dw)                                 & 288                                                                                                & 3612672                                                                                            & 69.10                  & 0.00     \\
Layer 3                                      & 2048                                                                                               & 25690112                                                                                           & 93.70                  & 90.00    \\
Layer 4 (dw)                                 & 576                                                                                                & 1806336                                                                                            & 62.50                  & 0.00     \\
Layer 5                                      & 8192                                                                                               & 25690112                                                                                           & 93.25                  & 90.00    \\
Layer 6 (dw)                                 & 1152                                                                                               & 3612672                                                                                            & 63.45                  & 0.00     \\
Layer 7                                      & 16384                                                                                              & 51380224                                                                                           & 92.19                  & 90.00    \\
Layer 8 (dw)                                 & 1152                                                                                               & 903168                                                                                             & 44.36                  & 0.00     \\
Layer 9                                      & 32768                                                                                              & 25690112                                                                                           & 94.65                  & 90.00    \\
Layer 10 (dw)                                & 2304                                                                                               & 1806336                                                                                            & 63.76                  & 0.00     \\
Layer 11                                     & 65536                                                                                              & 51380224                                                                                           & 94.91                  & 90.00    \\
Layer 12 (dw)                                & 2304                                                                                               & 451584                                                                                             & 53.43                  & 0.00     \\
Layer 13                                     & 131072                                                                                             & 25690112                                                                                           & 96.86                  & 90.00    \\
Layer 14 (dw)                                & 4608                                                                                               & 903168                                                                                             & 67.93                  & 0.00     \\
Layer 15                                     & 262144                                                                                             & 51380224                                                                                           & 97.25                  & 90.00    \\
Layer 16 (dw)                                & 4608                                                                                               & 903168                                                                                             & 78.43                  & 0.00     \\
Layer 17                                     & 262144                                                                                             & 51380224                                                                                           & 96.71                  & 90.00    \\
Layer 18 (dw)                                & 4608                                                                                               & 903168                                                                                             & 77.00                  & 0.00     \\
Layer 19                                     & 262144                                                                                             & 51380224                                                                                           & 95.40                  & 90.00    \\
Layer 20 (dw)                                & 4608                                                                                               & 903168                                                                                             & 74.22                  & 0.00     \\
Layer 21                                     & 262144                                                                                             & 51380224                                                                                           & 93.52                  & 90.00    \\
Layer 22 (dw)                                & 4608                                                                                               & 903168                                                                                             & 62.02                  & 0.00     \\
Layer 23                                     & 262144                                                                                             & 51380224                                                                                           & 90.64                  & 90.00    \\
Layer 24 (dw)                                & 4608                                                                                               & 225792                                                                                             & 53.78                  & 0.00     \\
Layer 25                                     & 524288                                                                                             & 25690112                                                                                           & 92.23                  & 90.00    \\
Layer 26 (dw)                                & 9216                                                                                               & 451584                                                                                             & 40.89                  & 0.00     \\
Layer 27                                     & 1048576                                                                                            & 51380224                                                                                           & 91.76                  & 90.00    \\
Layer 28 (fc)                                & 1024000                                                                                            & 1024000                                                                                            & 76.81                  & 90.00    \\ 
AP (average pool before fc) & 0 & 50176 & ~0.00 & ~0.00\\\bottomrule
\end{tabular}
}
\end{table}

\subsection{\alg Adaptations}
\label{sec:adapt}
Algorithm~\ref{alg:code} has comments suggesting the simple modifications required for global and per-weight sparsity.
\subsubsection{\alg for Global Sparsity}
\label{sec:global}
\alg can be trivially modified to learn the global threshold to induce global sparsity like in~\citep{han2015learning, frankle2018the}. Instead of having an $s_l$ per layer $l$, share all the $s_l$ to create one single learnable global threshold $s_{g}$. This can be implemented by a simple modification in Algorithm~\ref{alg:code}. \alg's capability to induce global sparsity was evaluated on ResNet50 for ImageNet-1K for 90-98\% sparsity regimes.
\begin{table}[!ht]
\centering
\caption{\alg can stablely learn the global threshold to induce global sparsity resulting in models with comparable accuracies as layer-wise sparsity but with $\sim2\times$ the inference cost.}
\label{tab:gs}
\resizebox{\columnwidth}{!}{
\begin{tabular}{@{}lcccr@{}}
\toprule
Method              & \begin{tabular}[c]{@{}c@{}}Top-1 Acc \\ (\%)\end{tabular} & Params & \begin{tabular}[c]{@{}c@{}}Sparsity \\ (\%)\end{tabular} & FLOPs \\ \midrule
\textit{\alg}-GS                 & 74.13                                                     & 2.42M  & 89.54                                                    & 596M  \\
\textit{\alg}-GS & {71.61}                                                     & {1.58M}  & {93.84}                                                     & {363M}  \\
\textit{\alg}-GS & {67.95}                                                     & {1.01M}  & {96.06}                                                     & {232M}  \\
\textit{\alg}-GS & 62.17                                                     & {0.54M}  & {97.91}                                                    &{142M}   \\ \bottomrule
\end{tabular}}
\end{table}

Table~\ref{tab:gs} shows the performance of \alg-GS that learns the global threshold to induce global sparsity. While the accuracies are comparable to the state-of-the-art if not better, they do come at cost of $\sim2\times$ inference cost compared to layer-wise sparsity due to poor non-uniform sparsity distribution which is a result of difference converged values of weights in each of the layers. \alg-GS has numbers similar to IMP~\citep{frankle2018the} while being able to learn the threshold stablely.

\subsubsection{\alg for Filter/Channel Pruning}
\label{sec:structured}
Let us assume there are $n_{out}$ filters of size $k\times k \times n_{in}$ in a given layer. Typically in channel/filter pruning techniques, each of these $n_{out}$ filters have an importance factor that represents the utility of the filter and is used to scale the corresponding filter. For a filter $f_i$ there exists a importance scalar $m_i$ learnt or obtained in some fashion and is used to get the effective filter in use $\hat{f}_i = m_i \cdot f_i$ where $m_i$ is broadcasted to scale $f_i$. In practice, $m_i$ is heuristically made to go to $0$ to induce structured sparsity through channel/filter pruning. Let us stack all the importance scalars of the filters in the layer, $\{m_i\}_{i\in[n_{out}]}$, as vector $\bm{m}_l$ where $l$ is the layer index. Now, this reduces to the same problem of inducing sparsity in a vector as in the learning of low-rank in RNN presented in Section~\ref{sec:RNN}. \alg will be applied to each of the $\{m_l\}_{l\in[L]}$ where $L$ is the total number of layers in a deep neural network. The inference will use the importance scalars through \alg ensuring channel/filter pruning due to the induced sparsity. This is very similar to the work-flow we used to induce low-rank in RNNs.

\subsubsection{\alg for Per-weight Pruning or Mask Learning}
The adaptation of \alg for per-weight pruning or mask learning is simple and is similar to layer-wise or global sparsity. Changing $s_l \to \mathbf{S}_l$ ie., changing the layer-wise thresholds from a scalar to a tensor of the size of $\mathbf{W}_l$ will hep STR adapt to do per-weight pruning or mask learning as discussed in the recent works~\citep{zhou2019deconstructing,savarese2019winning,ramanujan2020s}. We have explored this using a couple of experiments on CIFAR-10~\citep{krizhevsky2009learning} and ImageNet-1K. We observed that high amounts of sparsity were induced and the routine is very aggressive compared to other sparsification methods. For example, we were able to get 90\% accuracy on CIFAR-10 using ResNet18 at a staggering 99.63\% sparsity (270$\times$ lesser parameters than the dense model) which results in 41K parameters pushing it into very under parameterized regime. We suggest caution when running per-weight sparsity experiments with any method due to the high variance in the final accuracy.

\subsection{Hyperparameters for Reproducibility}
\label{sec:hyperparams}
\begin{table}[!ht]
\centering
\caption{The hyperparameters for various sparse ResNet50 models on ImageNet-1K using \alg. $\lambda$ is the weight-decay parameter and $s_\text{init}$ is the initialization of all $s_i$ for all the layers in ResNet50.}
\label{tab:res50hyperparams}
\resizebox{\columnwidth}{!}{
\begin{tabular}{@{}ccr@{}}
\toprule
Sparse Model (\%) & Weight-decay ($\lambda$) & \multicolumn{1}{c}{$s_{\text{init}}$} \\ \midrule
79.55             & 0.00001700000000        & -3200                       \\
81.27             & 0.00001751757813        & -3200                       \\
87.70             & 0.00002051757813        & -3200                       \\
90.23             & 0.00002251757813        & -3200                       \\
90.55             & 0.00002051757813        & -800                        \\
94.80             & 0.00003051757813        & -3200                       \\
95.03             & 0.00003351757813        & -12800                      \\
95.15             & 0.00003051757813        & -1600                       \\
96.11             & 0.00003051757813        & -100                        \\
96.53             & 0.00004051757813        & -12800                      \\
97.78             & 0.00005217578125        & -12800                      \\
98.05             & 0.00005651757813        & -12800                      \\
98.22             & 0.00006051757813        & -12800                      \\
98.79             & 0.00007551757813        & -12800                      \\
98.98             & 0.00008551757813        & -12800                      \\
99.10             & 0.00009051757813        & -12800                      \\ \bottomrule
\end{tabular}
}
\end{table}

All the ResNet50 experiments use a batchsize of 256, cosine learning rate with warm-up as in~\citep{wortsman2019discovering} and trained for 100 epochs. $\lambda$ is the weight-decay hyperparameter. $s_{\text{init}}$ is the initial value of all $s_i$ where $i$ is the layer number. The hyper parameter setting for each of the sparse model can be found in Table~\ref{tab:res50hyperparams}.

\begin{table}[!ht]
\centering
\caption{The hyperparameters for various sparse MobileNetV1 models on ImageNet-1K using \alg. $\lambda$ is the weight-decay parameter and $s_\text{init}$ is the initialization of all $s_i$ for all the layers in MobileNetV1.}
\label{tab:mbv1hyperparams}
\resizebox{\columnwidth}{!}{
\begin{tabular}{@{}ccr@{}}
\toprule
Sparse Model (\%) & Weight-decay ($\lambda$) & \multicolumn{1}{c}{$s_{\text{init}}$} \\ \midrule
75.28             & 0.00001551757813         & -100                                  \\
79.07             & 0.00001551757813         & -25                                   \\
85.80             & 0.00003051757813         & -3200                                 \\
89.01             & 0.00003751757813         & -12800                                \\
89.62             & 0.00003751757813         & -3200                                 \\ \bottomrule
\end{tabular}
}
\end{table}
All the MobileNetV1 experiments use a batchsize of 256, cosine learning rate with warm-up as in~\citep{wortsman2019discovering} and trained for 100 epochs. $\lambda$ is the weight-decay hyperparameter. $s_{\text{init}}$ is the initial value of all $s_i$ where $i$ is the layer number. The hyper parameter setting for each of the sparse model can be found in Table~\ref{tab:mbv1hyperparams}.

All the CNN experiments use $g(s) = \frac{1}{1+e^{-s}}$ for the \alg.

\begin{table}[!ht]
\centering
\caption{Hyperparameters for the low-rank FastGRNN with \alg. The same weight-decay parameter $\lambda$ is applied on both $\mathbf{m}_{\mathbf{W}}, \mathbf{m}_{\mathbf{U}}$. Multiple rank setting can be acheived during the training course of the FastGRNN model. $g(s_{\text{init}}) \approx 0$ ie., $s_{\text{init}} \le -10$ for all the experiments.}
\label{tab:rnnhyperparams}
\resizebox{\columnwidth}{!}{
\begin{tabular}{@{}cc|cc@{}}
\toprule
\multicolumn{2}{c}{Google-12}       & \multicolumn{2}{|c}{HAR-2}          \\ \midrule
($r_W$, $r_U$) & Weight-decay ($\lambda$) & ($r_W$, $r_U$) & Weight-decay ($\lambda$) \\ \midrule
(12, 40) & 0.001                    & (9, 8)   & 0.001                    \\
(11, 35) & 0.001                    & (9, 7)   & 0.001                    \\
(10, 31) & 0.002                    & (8, 7)   & 0.001                    \\
(9, 24)  & 0.005                    &          &                          \\ \bottomrule
\end{tabular}
}
\end{table}
All the FastGRNN experiments use a batchsize of 100, learning rate and optimizers as suggested in~\citep{kusupati2018fastgrnn} and trained for 300 epochs. Weight-decay parameter, $\lambda$ is applied to both  $\mathbf{m}_{\mathbf{W}}, \mathbf{m}_{\mathbf{U}}$ resulting in the rank setting obtained. Each hyperparamter setting can lead to multiple low-rank setting over the course of training. $s_{\mathrm{init}}$ set such that $g(s_{\mathrm{init}}) \approx 0$ for the initialization of soft threshold pruning scalar for the low-rank vectors.

All the RNN experiments use $g(s) = e^s$ for the \alg.

\subsection{Dataset and Model Details}
\label{sec:datasetdet}
\textbf{ImageNet-1K:} ImageNet-1K has RGB images with 224$\times$224 dimensions. The dataset has 1.3M training images, 50K validation images and 1000 classes. Images were transformed and augmented with the standard procedures as in~\citep{wortsman2019discovering}.

\textbf{Google-12:} Google Speech Commands dataset~\citep{warden2017} contains 1 second long utterances of 30 short words (30 classes) sampled at 16KHz. Standard log Mel-filter-bank featurization with 32 filters over a window size of 25ms and stride of 10ms gave 99 timesteps of 32 filter responses for a 1-second audio clip. For the 12 class version, 10 classes used in Kaggle’s Tensorflow Speech Recognition challenge were used and the remaining two classes were noise and background sounds (taken randomly from the remaining 20 short word utterances). The datasets were zero mean - unit variance normalized during training and prediction. Google-12 has 22,246 training points, 3,081 testing points. Each datapoint has 99 timesteps with each input being 32 dimensional making the datapoint 3,168 dimensional.

\textbf{HAR-2:} Human Activity Recognition (HAR) dataset was collected from an accelerometer and gyroscope on a Samsung Galaxy S3 smartphone. The features available on the repository were directly used for experiments. The 6 activities were merged to get the binarized version. The classes \{Sitting, Laying, Walking\_Upstairs\} and \{Standing, Walking, Walking\_Downstairs\} were merged to obtain the two classes. The dataset was zero mean - unit variance normalized during training and prediction. HAR-2 has 7,352 training points and 2,947 test points. Each datapoint has 1,152 dimensions, which will be split into 128 timesteps leading to dimensional per timestep inputs.

\textbf{ResNet50:} ResNet50 is a very popular CNN architecture and is widely used to showcase the effectiveness of sparsification techniques. ResNet50 has 54 parameter layers (including fc) and a couple of pooling layers (which contribute minimally to FLOPs). All the batchnorm parameters are left dense and are learnt during the training. \alg can be applied per-layer, per-channel and even per-weight to obtain unstructured sparsity and the aggressiveness of sparsification increases in the same order. This paper only uses per-layer \alg which makes it have 54 additional learnable scalars. The layer-wise parameters and FLOPs can be seen in Tables~\ref{tab:res50nub} and~\ref{tab:res50big}. All the layers had no bias terms.

\textbf{MobileNetV1:} MobileNetV1 is a popular efficient CNN architecture. It is used to showcase the generalizability of sparsification techniques. MobileNetV1 has 28 parameter layers (including fc) and a couple of pooling layers (which contribute minimally to FLOPs). All the batchnorm parameters are left dense and are learnt during the training. \alg can be applied per-layer, per-channel and even per-weight to obtain unstructured sparsity and the aggressiveness of sparsification increases in the same order. This paper only uses per-layer \alg which makes it have 28 additional learnable scalars. The layer-wise parameters and FLOPs can be seen in Tables~\ref{tab:mbv1nub}. All the layers had no bias terms.

\textbf{FastGRNN:} FastGRNN's update equations can be found in~\citep{kusupati2018fastgrnn}. FastGRNN, in general, benefits a lot from the low-rank reparameterization and this enables it to be deployed on tiny devices without losing any accuracy. FastGRNN's biases and final classifier are left untouched in all the experiments and only the input and hidden projection matrices are made low-rank. All the hyperparameters were set specific to the datasets as in~\citet{kusupati2018fastgrnn}.

\subsection{Hard Threshold vs Soft Threshold}
\begin{figure}[!ht]
	\includegraphics[width=\columnwidth]{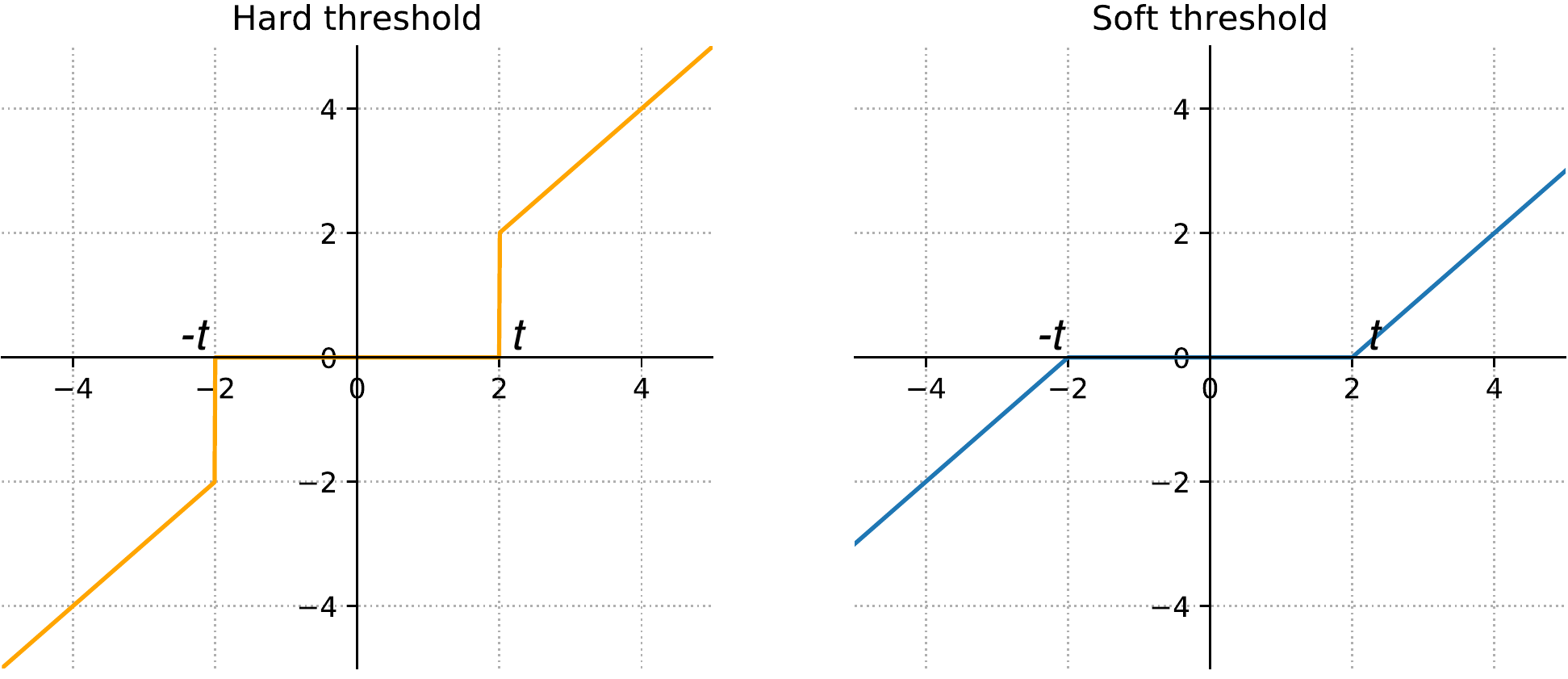}
	\caption{A visualization of hard-threshold (\textit{left}) and soft-threshold (\textit{right}) functions with the threshold $\alpha = 2$. x-axis is the input and y-axis is the output.}
 	\label{fig:stht}
\end{figure}

Figure~\ref{fig:stht} shows the difference between hard thresholding and soft thresholding for the same threshold value of $\alpha = 2$. It is clear from Figure~\ref{fig:stht} that soft-threshold is a continuous function that is sub-differentiable. The abrupt change in hard-threshold leads to instability in training sometimes increasing dependence on fine tuning of the obtained sparse network. Soft-threshold is robust to such issues. 

\end{document}